\documentclass[10pt,twocolumn,letterpaper]{article}

\usepackage{cvpr}              % To produce the CAMERA-READY version
\definecolor{cvprblue}{rgb}{0.21,0.49,0.74}
\usepackage[pagebackref,breaklinks,colorlinks,allcolors=cvprblue]{hyperref}
\usepackage{algorithmic}
\usepackage{algorithm}
\usepackage{bbding}
\usepackage{color, colortbl}
\usepackage{multirow}
\def\paperID{256} % *** Enter the Paper ID here
\def\confName{CVPR}
\def\confYear{2025}

\title{DoF-Gaussian: Controllable Depth-of-Field for 3D Gaussian Splatting}

\author{Liao Shen$^{1,2}$\hspace{0.1in} 
        Tianqi Liu$^{1,2}$\hspace{0.1in} 
        Huiqiang Sun$^{1,2}$\hspace{0.1in} 
        Jiaqi Li$^{1}$\hspace{0.1in} 
        Zhiguo Cao$^{1}$\hspace{0.1in}
        Wei Li$^{2}$\footnotemark[1]~\hspace{0.1in}
        Chen Change Loy$^{2}$\hspace{0.1in}\\
$^1$School of AIA, Huazhong University of Science and Technology\\
$^2$S-Lab, Nanyang Technological University\hspace{0.3in}\\
{\tt\small leoshen@hust.edu.cn}}
\begin{document}
%\maketitle
\twocolumn[{%
\renewcommand\twocolumn[1][]{#1}%
\maketitle
\vspace{-10mm}
\begin{center}
    \centering
    \includegraphics[width=\linewidth]{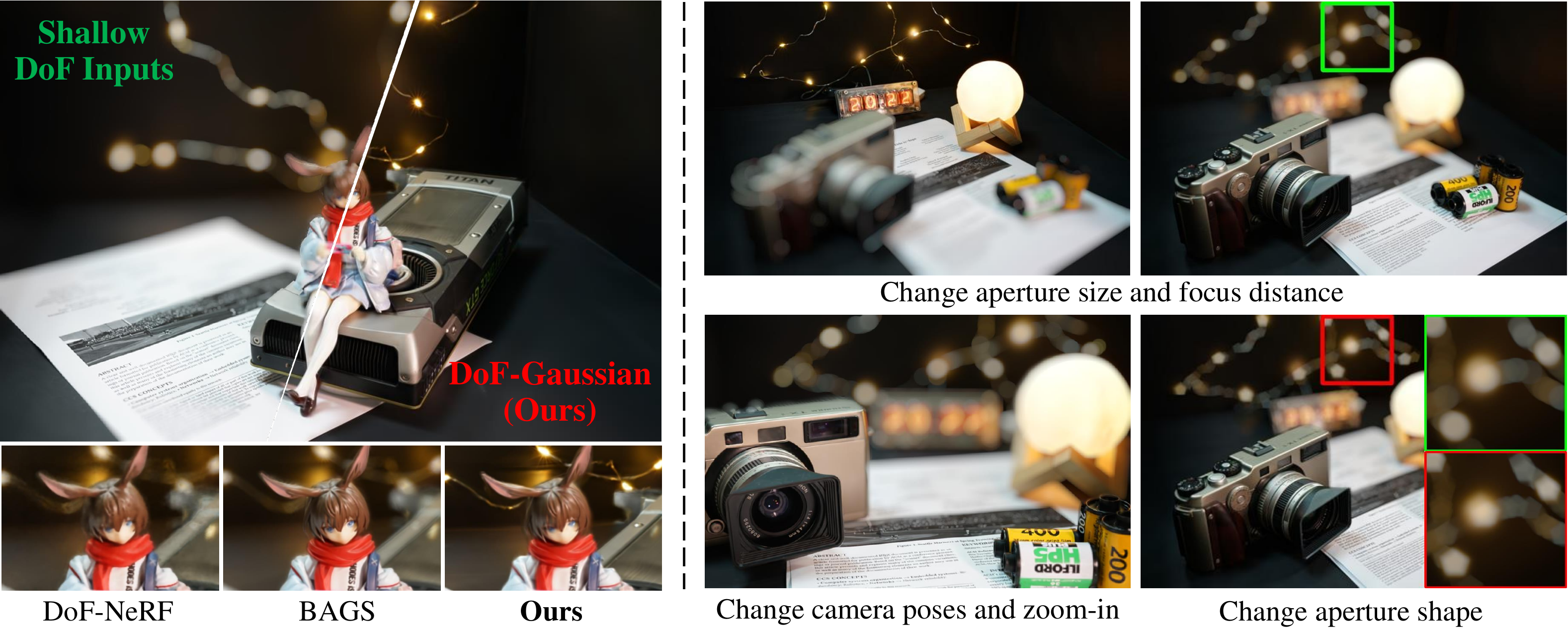}
    \vspace{-6mm}
    \captionof{figure}{
    Given a set of multi-view input images with shallow DoF, DoF-Gaussian can reconstruct a 3D-GS representation of a sharp scene. Thanks to our lens-based design, we can also achieve controllable DoF effects for a variety of applications. The example input images are taken from~\cite{wu2022dof} for illustration purposes. \textbf{(Zoom-in for best view)}
    } 
    \label{fig:teaser}
\end{center}%
}]
\renewcommand{\thefootnote}{\fnsymbol{footnote}} %将脚注符号设置为fnsymbol类型，即特殊符号表示
\footnotetext[1]{Corresponding author.}
\begin{abstract}
Recent advances in 3D Gaussian Splatting (3D-GS) have shown remarkable success in representing 3D scenes and generating high-quality, novel views in real-time. However, 3D-GS and its variants assume that input images are captured based on pinhole imaging and are fully in focus. This assumption limits their applicability, as real-world images often feature shallow depth-of-field (DoF). In this paper, we introduce DoF-Gaussian, a controllable depth-of-field method for 3D-GS. We develop a lens-based imaging model based on geometric optics principles to control DoF effects. To ensure accurate scene geometry, we incorporate depth priors adjusted per scene, and we apply defocus-to-focus adaptation to minimize the gap in the circle of confusion. We also introduce a synthetic dataset to assess refocusing capabilities and the model’s ability to learn precise lens parameters. Our framework is customizable and supports various interactive applications. Extensive experiments confirm the effectiveness of our method. 
Our project is available at \href{https://dof-gaussian.github.io/}{https://dof-gaussian.github.io/}.

\end{abstract}    
\section{Introduction}
\label{sec:intro}
Depth-of-field (DoF) refers to the distance between the closest and farthest objects in a photo that appears acceptably sharp. In practice, photographers can control the DoF by adjusting the camera's aperture size or focus distance to capture images with either a wide DoF (all-in-focus) or shallow DoF (defocused).
The shallow DoF effect is an important technique in photography, as it draws the viewers’ attention to the focal region by blurring the surrounding areas. Meanwhile, novel view synthesis aims to create realistic images from novel viewpoints based on a set of source images. However, novel view synthesis typically requires all-in-focus input images and lacks the capability to render varied DoF effects, limiting its applications. In this work, we aim to render novel views with controllable DoF, adding cinematic quality to the results.

Images captured from the real world sometimes have shallow DoF. Specifically, points of light that do not lie on the focal plane are projected onto the sensor plane as blurred circles, referred to as the circle of confusion (CoC), and this introduces bokeh blur to the captured image.
Most novel view synthesis methods experience performance degradation when processing shallow DoF input images. To address this issue, DoF-NeRF~\cite{wu2022dof} and LensNeRF~\cite{kim2024lensnerf} introduce lens-based camera models in volume rendering to explicitly enable controllable DoF effects. Meanwhile, methods like Deblur-NeRF~\cite{ma2022deblur}, DP-NeRF~\cite{lee2023dp}, and PDRF~\cite{peng2023pdrf} employ sparse kernels to model defocus blur. However, 
implicit rendering approaches face significant challenges in training and rendering efficiency.
In the realm of 3D Gaussian Splatting (3D-GS)~\cite{kerbl20233d}, methods such as BAGS~\cite{peng2024bags} and Deblurring 3DGS~\cite{lee2024deblurring} propose blur estimation networks to model the blur kernels or scaling factors for defocus deblurring, but they cannot accommodate controllable DoF. 

\if 0
Due to the limitation of existing methods for shallow DoF inputs, in this paper, we aim to develop an efficient controllable depth-of-field framework for 3D Gaussian Splatting, which faces several critical challenges: 
\textbf{\textit{First}}, 3D-GS and its variants~\cite{yu2024mip,liu2025mvsgaussian,huang20242d,wu20244d} are generally based on the pinhole camera model and assume all-in-focus inputs, i.e., both foreground and background are clear. 
\textbf{\textit{Second}}, when the input images have bokeh blur, it is hard to construct scene geometry and render depth map correctly. However, rendering images with different depths of field relies on the correct depth. 
\textbf{\textit{Third}}, the algorithm assumes that the CoC is ideal and therefore will differ from the CoC in real photos. This inconsistency will hinder the defocus deblurring results.
\textbf{\textit{Last}}, existing dataset in this domain are mainly borrowed from the deblurring domain. We can only evaluate the model’s defocus deblurring ability, but ignore other capacity of controllable DoF.
\fi

In this paper, we present DoF-Gaussian, an efficient framework for controllable DoF in 3D-GS, addressing the limitations of existing methods in handling shallow DoF inputs. This problem is non-trivial because 3D-GS and its variants~\cite{yu2024mip,liu2025mvsgaussian,huang20242d,wu20244d,xu2024depthsplat} are typically based on a pinhole camera model that assumes all-in-focus inputs, meaning both foreground and background appear clear.
When input images contain bokeh blur, it becomes challenging to accurately construct scene geometry and render depth maps. However, rendering images with different DoFs relies heavily on precise depth estimation.
Additionally, our 3D-GS approach assumes an idealized CoC, which inevitably differs from the CoC seen in real photographs, posing further challenges for accurate defocus deblurring.
Furthermore, existing datasets in this field are primarily borrowed from deblurring applications, limiting our evaluation to the model's defocus deblurring capability and overlooking other aspects of controllable DoF effects.

\if 0
To tackle these challenges point by point:  
\textbf{\textit{First}}, we simulate the lens-based model rather than pinhole-based model. Meanwhile, we set the lens optical parameters aperture size and focus distance to learnable in order to control the depth of field.  
\textbf{\textit{Second}}, we present the per-scene adjustment of depth priors to minimize scene geometry and depth relation errors, which helps to correctly control the depth of field.  
\textbf{\textit{Third}}, we propose a defocus-to-focus adaptation strategy to focus on learning the focal region after the learnable lens parameters converge, as a way to compensate for the difference in CoC.
\textbf{\textit{Last}}, we introduce a synthetic dataset used to comprehensively evaluate the model, including the ability to refocus and verify that our model learns the correct lens parameters.
\fi

To address these challenges, we present a new controllable DoF 3D-GS method with the following \textbf{contributions}:
\textbf{\textit{First}}, we employ a lens-based model instead of a pinhole-based one. We make the lens optical parameters, such as aperture size and focus distance, learnable to enable control over the depth of field.
\textbf{\textit{Second}}, we introduce per-scene adjustments for depth priors to minimize errors in scene geometry and depth relationships, ensuring accurate depth-of-field control.
\textbf{\textit{Third}}, we propose a defocus-to-focus adaptation strategy that focuses on learning the focal region after the learnable lens parameters have converged, compensating for differences in CoC.
\textbf{\textit{Finally}}, we present a synthetic dataset designed to comprehensively evaluate the model, including its refocusing capabilities and its ability to learn accurate aperture size and focus distance.

% \begin{figure}[!t]
%   \centering
%   \setlength{\belowcaptionskip}{-5pt}
%   \includegraphics[width=\linewidth]{figs/coc_size.pdf}
%   \caption{\normalsize{Users can interactively adjust the shape of the CoC, similar to how switching camera lenses works.}
%   }
%   \label{fig:coc}
% \end{figure}

Benefiting from the lens-based imaging model, our approach enables controlled depth of field, allowing it to handle not only shallow DoF image inputs but also unlock a range of interactive and engaging applications, as depicted in Figure~\ref{fig:teaser}. For instance, users can render novel view images with varied DoF, refocus on custom datasets, and even adjust the shape of CoC, \eg, from circular to pentagonal or hexagonal CoC
. Moreover, users can dynamically change the DoF while moving the camera or zooming,  rendering videos with cinematic effects. Extensive experiments demonstrate that our method outperforms state-of-the-art depth-of-field and defocus deblurring methods.

\if 0
In summary, we present a new controllable DoF DS-GS method.
Extensive experiments demonstrate that our method outperforms state-of-the-art depth-of-field and defocus deblurring methods. Our main contributions can be summarized as follows:
\begin{itemize}[leftmargin=*]
    \item We present DoF-Gaussian, a controllable depth-of-field 3D Gaussian Splatting derived from a lens-based model.
    \item We employ a per-scene adjustment of depth priors to guarantee the correct scene geometry and depth map.
    \item We further propose a defocus-to-focus adaptation strategy to bridge the gap in CoC.
    \item We introduce a synthetic dataset for a more comprehensive evaluation of the model.
\end{itemize}
\fi

% To insert a figure: \input{figs/template}
% Or table: \input{tables/template}

\section{Related Work}
\label{sec:related}

\label{sec:method}
\begin{figure*}[!t]
  \centering
  \setlength{\belowcaptionskip}{-5pt}
  \includegraphics[width=\linewidth]{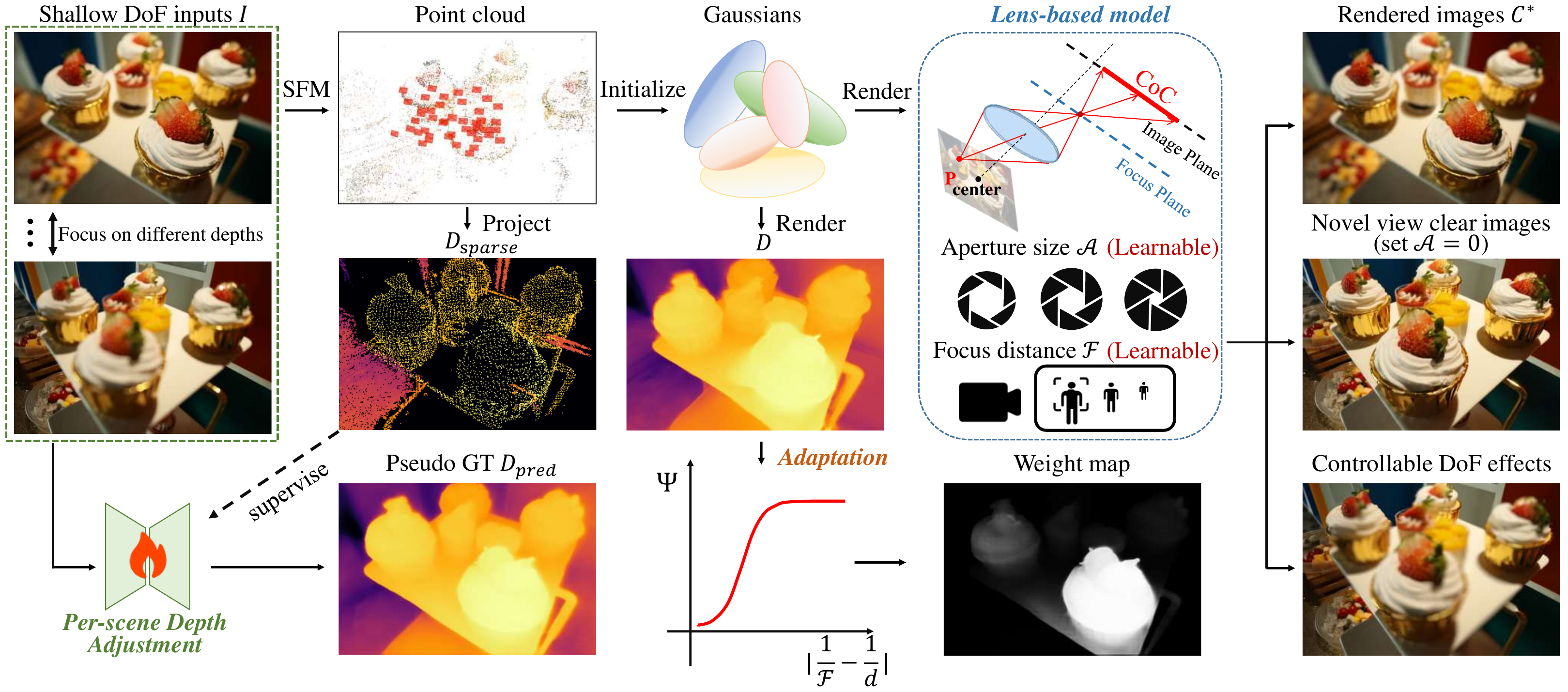}
  \caption{\textbf{Overview of DoF-Gaussian.} Given input images $I$ with shallow DoF , we first apply SfM from COLMAP to obtain sparse depth $D_{sparse}$, which is used to train a depth network to derive per-scene depth priors $D_{pred}$. We then employ $D_{pred}$ to regularize the Gaussians rendered depth map $D$. Next, by developing a lens imaging model, we can render defocused images $C^*$ to simulate input images. To minimize the discrepancy in CoC, we propose an adaptation using the weight map. Finally, we can render fully clear images for novel view synthesis and achieve various effects by our controllable DoF framework.
  }
  \label{fig:pipeline}
\end{figure*}

\noindent\textbf{Depth-of-Field Rendering.}
Rendering DoF effects from a single all-in-focus image has been widely explored in previous work. Bokeh blur occurs when light is projected onto the image plane as a circular region, rather than as a point.
The size of CoC is affected by the diameter of the aperture (aperture size) and the distance from the camera to the focal plane (focus distance). Photos captured with a small aperture usually present a wide DoF, \ie, all objects appear sharp. In contrast, as the aperture diameter increases, objects near the focal plane remain sharp, while those farther away become blurred with a larger CoC.
Physically based methods~\cite{abadie2018advances, lee2010real} rely on 3D scene geometry information and are time-consuming. Some methods use neural networks~\cite{ignatov2020rendering, dutta2021stacked, qian2020bggan, ignatov2020aim, ignatov2019aim} trained end-to-end to obtain shallow DoF images. The DoF effects in some studies~\cite{peng2022bokehme, conde2023lens, xiao2018deepfocus, wadhwa2018synthetic,wang2018deeplens, sheng2024dr, zhang2019synthetic, luo2024video,peng2023selective, piche2023lens} are controllable but usually require an extra disparity map. Wang \etal~\cite{wang2023implicit} combines neural fields with an expressive camera model to achieve all-in-focus reconstruction from an image stack. In this work, we show the connection between DoF rendering and novel view synthesis using a lens-based camera model.

\noindent\textbf{Image Deblurring.}
Blur can generally be categorized into two main types: camera motion blur and defocus blur. Previous studies~\cite{ma2022deblur,peng2023pdrf,peng2024bags,darmon2024robust,lee2024deblurring,sun2024dyblurf,luo2024dynamic} attempt to address this issue.
Our task of rendering all-in-focus and sharp novel view images from shallow DoF inputs can be considered as a form of defocus deblurring. By modeling the underlying physical principles, we can achieve defocus deblurring by simulating the circle-of-confusion. Furthermore, our method allows for adjustments to lens parameters, enabling applications such as refocusing—capabilities that deblurring approaches like BAGS~\cite{peng2024bags} cannot achieve.
Existing datasets in the 3D DoF field are often borrowed from the deblurring domain~\cite{ma2022deblur} or designed solely to assess defocus deblurring ability~\cite{wu2022dof}. To support a wider range of applications and validate the accuracy of the learned lens model, we introduce a synthetic dataset for more comprehensive evaluations.

\noindent\textbf{Novel View Synthesis.}
Novel view synthesis allows the rendering of unseen camera perspectives from 2D images and their corresponding camera poses. Recent advancements in synthesis can largely be attributed to Neural Radiance Field~\cite{mildenhall2021nerf} and 3D Gaussian Splatting~\cite{kerbl20233d}. 
However, real-world images are often not clean and well-calibrated. Previous work~\cite{wang2023bad,zhao2024bad,chen2024deblur,mildenhall2022nerf,martin2021nerf,park2021nerfies,sun2024dyblurf,wu20244d,shen2023make} attempts to cope with non-ideal conditions, such as low light, camera motion, bokeh effects, and various types of image degradation. 
To recover a sharp scene from images with bokeh blur, methods like Deblur-NeRF~\cite{ma2022deblur}, DP-NeRF~\cite{lee2023dp}, and PDRF~\cite{peng2023pdrf} model defocus blur using sparse kernels. NeRFocus~\cite{wang2022nerfocus} and LensNeRF~\cite{kim2024lensnerf} achieve defocus effects by incorporating a lens-based camera model in volume rendering. 
While DoF-NeRF~\cite{wu2022dof} is similar to our approach, it is limited by the training and rendering efficiency and quality of NeRF. BAGS~\cite{peng2024bags} and Deblurring 3DGS~\cite{lee2024deblurring} use blur estimation networks to model blur kernels or scaling factors but cannot achieve controllable DoF effects due to the absence of a lens imaging model. Other studies, such as
\cite{wang2024cinematic,jin2024lighting}, focus on HDR tasks and consider depth-of-field simultaneously.  RGS~\cite{darmon2024robust} addresses defocus blur by introducing an offset to the 2D covariance matrices of Gaussians.~\cite{moenne20243d} introduces a ray tracing rendering algorithm for particle-based representations to enable many advanced techniques, such as the shallow depth-of-field effect.
A concurrent work~\cite{wang2024dof} also explores adjustable depth of field. Our method differs in its controllable DoF mechanism, with the introduction of depth priors to ensure accurate scene depth. In addition, our new defocus-to-focus adaptation strategy eliminates the need for the focus localization network proposed in their work. Besides, we introduce a synthetic dataset for comprehensive evaluation.
\section{Preliminary}
\textbf{3D Gaussian Splatting} represents a 3D scene as a mixture of anisotropic 3D Gaussians, where each Gaussian is characterized by a 3D covariance matrix $\mathit{\Sigma}$ and mean $\mu$:
\begin{equation}
    G(X) = e^{-\frac{1}{2}(X-\mu)^\mathsf{T}\mathit{\Sigma}^{-1}(X-\mu)} \tag{1} \label{eq1}
\end{equation}
The covariance matrix $\mathit{\Sigma}$ holds physical meaning only when it is positive semi-definite. Therefore, to enable effective optimization via gradient descent, $\mathit{\Sigma}$ is decomposed into a scaling matrix $S$ and a rotation matrix $R$, as: 
\begin{equation}
    \mathit{\Sigma} = RSS^\mathsf{T}R^\mathsf{T}. \tag{2} \label{eq2}
\end{equation}

To splat Gaussians from 3D space to a 2D plane, the view transformation matrix $W$ and the Jacobian matrix $J$, which represents the affine approximation of the projective transformation, are utilized to obtain the covariance matrix $\mathit{\Sigma}'$ in 2D space, as:
\begin{equation}
    \mathit{\Sigma}'=JW \mathit{\Sigma} W^\mathsf{T}J^\mathsf{T}. \tag{3} \label{eq3}
\end{equation}

Subsequently, a point-based alpha-blending rendering can be performed to determine the color of each pixel:
\begin{equation}
 C=\sum_i c_i\alpha_i \prod_{j=1}^{i-1}(1-\alpha_i), \tag{4} \label{eq4}
\end{equation}
 where $c_i$ represents the color of each point, defined by spherical harmonics (SH) coefficients. 
The density $\alpha_i$ is computed as the product of 2D Gaussians and a learnable point-wise opacity. 
% During optimization, the learnable attributes of each Gaussian are updated through gradient descent, including 1) a 3D position  $\mu \in \mathbb{R}^3$, 2) a scaling vector $s \in \mathbb{R}^3_+$,
%  3) a quaternion rotation vector $r \in \mathbb{R}^4$, 4) a color defined by SH $c \in \mathbb{R}^k$ (where k is the freedom), and 5) an opacity $\alpha \in [0,1]$. 
%  Additionally, an adaptive density control module is introduced to improve rendering quality, comprising mainly the following three operations: 1) split into smaller Gaussians if the magnitude of the scaling exceeds a threshold, 2) clone if the magnitude of the scaling is smaller than a threshold, and 3) prune
%  Gaussians with excessively small opacity or overly large scaling magnitudes.
\section{Method}
We propose \textit{DoF-Gaussian}, a controllable DoF approach for 3D-GS. The overall pipeline is illustrated in Fig.~\ref{fig:pipeline}. In Section~\ref{sec:4.1}, we first develop a lens-based imaging model based on geometric optics principles to control DoF effects. Next, we employ the per-scene adjustment of depth priors to guide the correct scene geometry, as described in Section~\ref{sec:4.2}.
Finally, in Section~\ref{sec:4.3}, we apply defocus-to-focus adaptation to minimize the inconsistencies in the CoC and enhance the defocus deblurring performance.

\subsection{Lens-based Depth-of-Field Model}
\label{sec:4.1}
\begin{figure}[!t]
  \centering
  \setlength{\belowcaptionskip}{-5pt}
  \includegraphics[width=\linewidth]{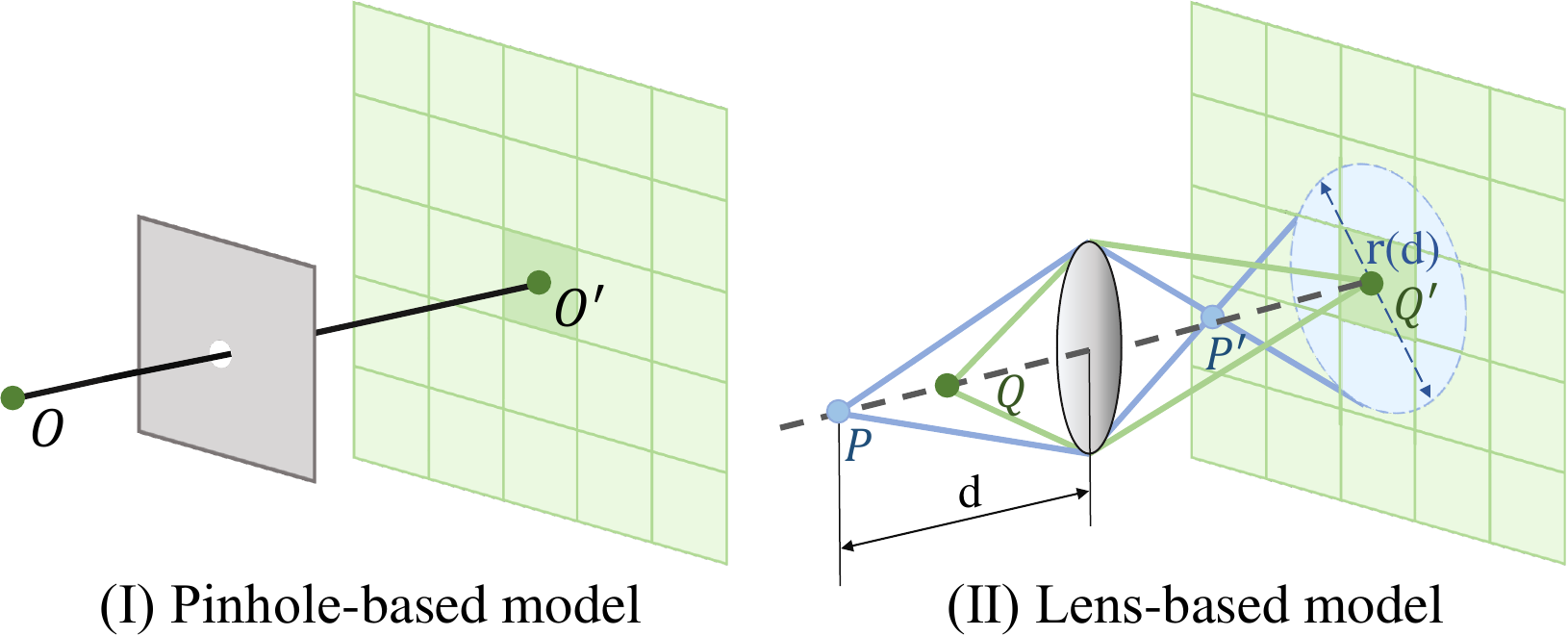}
  \caption{\textbf{The difference between the pinhole-based model ($\mathrm{I}$) and the lens-based model ($\mathrm{II}$).} For ($\mathrm{I}$), light emitted from spatial point $O$ directly hits the point $O'$ of the image plane. For ($\mathrm{II}$), we show the case that light emitted from $Q$ converges on the point $Q'$ of the image plane (\textbf{In-Focus}), and the case that light emitted from $P$ converges on the point $P'$ and continues to scatter onto the image plane forming a circle of confusion (\textbf{Out-of-Focus}). 
  }
  \label{fig:lens}
\end{figure}
The physical principles underlying imaging and DoF have been extensively studied in the field of geometric optics~\cite{hecht2012optics,pedrotti2018introduction}.
In an idealized optical system, the light emitted from a spatial point $O$ is projected onto a corresponding point $O'$ in the image plane following the principles of pinhole imaging, as illustrated in Fig.~\ref{fig:lens} ($\mathrm{I}$).
However, real-world cameras operate based on the lens model, as shown in Fig.~\ref{fig:lens} ($\mathrm{II}$). We show the two cases of in-focus and out-of-focus imaging, respectively. A spatial point $P$, located at a distance $d$ from the lens, is projected on the imaging plane as a circular region referred to as the circle of confusion (CoC). The diameter of this region, $r(d)$, can be determined by the aperture parameter $\mathcal{A}$ and focus distance $\mathcal{F}$, according to the following equation:

\begin{equation}
    r(d)=\mathcal{A}\bigg|\frac{1}{\mathcal{F}}-\frac{1}{d}\bigg|.  \tag{5} \label{eq5}
\end{equation}

The focus distance $\mathcal{F}$ primarily controls the depth position of the focal plane of the image, \ie, the depth of the sharp area, while the aperture parameter $\mathcal{A}$ determines the extent of the bokeh effect. 
For the point $Q$ located at the focus distance $d=\mathcal{F}$, the emitted rays directly converge through the lens to the corresponding point $Q'$ in the image plane, resulting in the absence of a circle of confusion (CoC). This also implies that $r(\mathcal{F})=0$.

We set focus distance $\mathcal{F}$ and aperture parameter $\mathcal{A}$ as learnable for each input image, and these parameters are continuously updated with the optimization of 3D-GS. 
By modeling the lens camera, the output $C^*$, which includes bokeh blur effects, can be derived from the 3D-GS rasterizer rendering result $C$ using our CUDA-based Algorithm~\ref{alg1}. 

Ideally, the confuse function is represented by the indicator function $\mathbb{I}(r(d)>l)$. To achieve a smooth and natural DoF effect, we substitute it with a differentiable function, as proposed in Busam~\etal~\cite{busam2019sterefo}:
\begin{equation}
    Func(d,l)=\frac{1}{2}+\frac{1}{2} \tanh\Big(\alpha \big(r(d)-l \big) \Big), \tag{6} \label{eq6}
\end{equation}
\begin{algorithm}[t]
\caption{Lens-based imaging process} 
\label{alg1} 
\renewcommand{\algorithmicrequire}{\textbf{Input:}}
\renewcommand{\algorithmicensure}{\textbf{Output:}}
\begin{algorithmic}
\REQUIRE Rasterization rendering $C$, aperture parameter $\mathcal{A}$, focus distance $\mathcal{F}$, depth map $D$, gamma value $\gamma$, confuse function $Func$
\ENSURE Defocus simulated result $C^*$
        \STATE $R \gets \mathcal{A}|\frac{1}{\mathcal{F}}-\frac{1}{D}|$ \quad $C \gets (C)^{\gamma}$
        \STATE $\Phi=[0], C^*=[0]$
        \FOR {pixel $i \gets \mathrm{Traverse}(C)$}
            \FOR {pixel $j \gets \mathrm{TraverseNeighbor}(c_i,r_i)$}
                \STATE $\lambda_{ij} \gets \mathrm{Func}(d_i,|i-j|)$
                \STATE $\Phi_j \gets \Phi_j+\lambda_{ij}$
                \STATE $c_j^* \gets c_j^*+\lambda_{ij} \cdot c_i $
            \ENDFOR
        \ENDFOR
        \STATE $C^* \gets (C^*/\Phi)^{\frac{1}{\gamma}}$
\end{algorithmic}
\end{algorithm}
where $\alpha$ defines the smoothness of confuse transition and $l$ represents the distance between two pixels. 

We supervise the training by computing the reconstruction loss between the output image $C^*$ and the shallow DoF input image $I$, where the loss function is a combination of the $\mathcal{L}_1$ and a D-SSIM term, with $\lambda$ set to $0.2$:
\begin{equation}
  \mathcal{L}_{rec}=(1-\lambda)\mathcal{L}_1(I, C^*) + \lambda \mathcal{L}_{D-SSIM}(I, C^*). \tag{7} \label{eq7}
\end{equation}

For inference, we set the aperture size $\mathcal{A}$ to $0$ so that we can render fully clear images for novel view synthesis.

\subsection{Per-Scene Adjustment of Depth Priors}
\label{sec:4.2}
Unlike previous work that assumes input images are all-in-focus and fully clear, it becomes challenging to accurately reconstruct scene geometry and render depth maps for bokeh blurring input images. However, rendering images with varying depths of field relies on precise depth information.
% Previous work primarily assumes that the input images are all-in-focus and fully clear. However, when the input images exhibit bokeh blur, accurately constructing scene geometry and rendering the depth map becomes challenging. Rendering images with varying depths of field relies on precise depth information.
\begin{figure}[!t]
  \centering
  \setlength{\belowcaptionskip}{-5pt}
  \includegraphics[width=\linewidth]{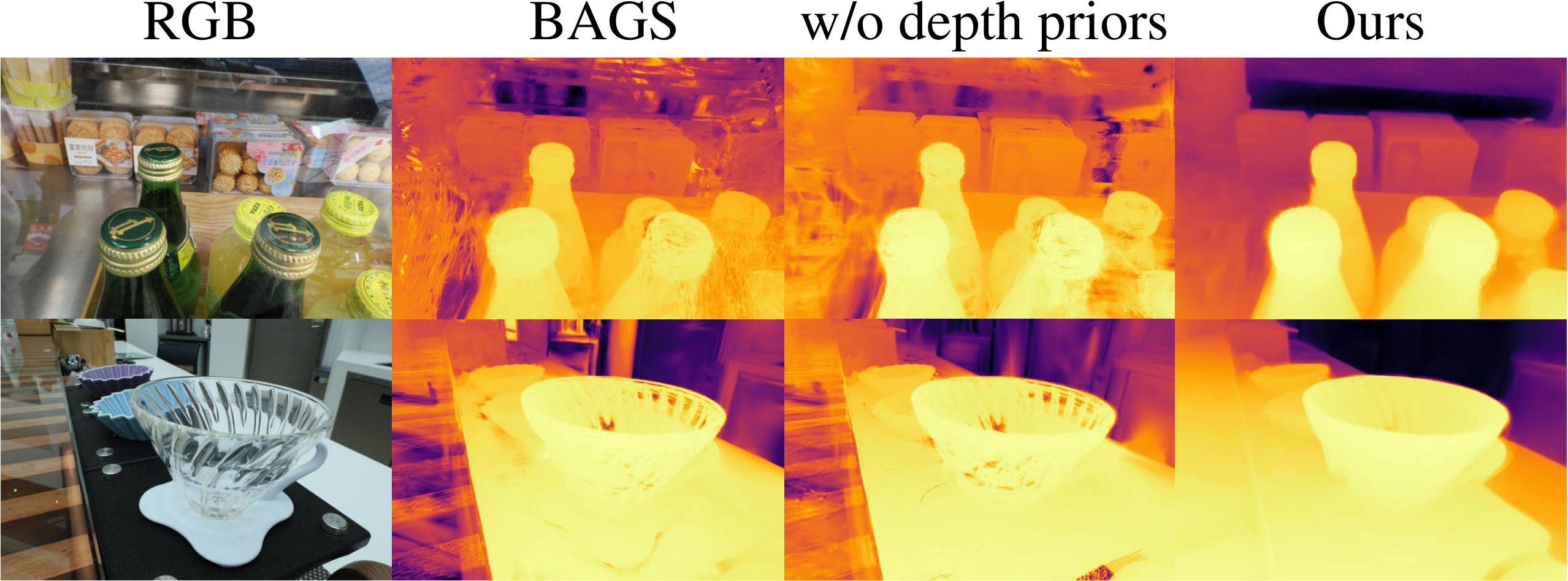}
  \caption{\textbf{Effects of the depth priors.} The per-scene adjustment of depth priors enhance the geometric structure and yield a more accurate depth map than that estimated by BAGS~\cite{peng2024bags} and a variant of our method without depth priors.
  }
  \label{fig:depth}
\end{figure}
To address this issue, we propose using depth prior as useful guidance for the accurate reconstruction of scene geometry. However, due to the presence of bokeh blur in shallow DoF inputs, directly employing a monocular depth network to predict depth maps $D_{pred}$ as pseudo ground truth to regularize the Gaussians rendered depth map, as done in~\cite{chung2024depth, li2024dngaussian}, does not yield satisfactory results. 
Inspired by~\cite{wei2021nerfingmvs}, we propose adapting per-scene depth priors through fine-tuning a monocular depth network based on the scene's sparse reconstruction. 
% We address this problem by adapting per-scene depth priors through fine-tuning a monocular depth network on the scene's sparse reconstruction. 
Specifically, we utilize COLMAP~\cite{schonberger2016structure, schonberger2016pixelwise} to obtain a sparse Structure-from-Motion (SfM) point cloud and create a set of 3D Gaussians. Consequently, we can derive per-view sparse depth maps $D_{sparse}$ by projecting the 3D point clouds after multi-view stereo processing. Although the resulting depth map is sparse, it is robust and can serve as a supervisory signal for training the per-scene depth priors. To address the scale ambiguity inherent in acquired depth maps, we apply a silog loss~\cite{eigen2014depth} to fine-tune the depth network~\cite{ranftl2020towards,li2019learning}, expressed as:
\begin{equation}
   \mathcal{L}_{silog} = \frac{1}{2M} \sum_{i=1}^M(\log(e^sD_{pred})-\log(D_{sparse}))^2, \tag{8} \label{eq8}
\end{equation}
where $e^s$ is the scale factor and $M$ is the number of pixels in the image used for calculation.
We apply the rasterization technique from Rade-GS~\cite{zhang2024rade}, which enables the robust rendering of depth maps $D$ for 3D-GS scenes. We predict the depth maps $D_{pred}$ from multi-view input images using the fine-tuned depth network and use them as depth priors to regularize $D$.  The depth loss can be described as follows:
\begin{equation}
    \mathcal{L}_{depth}=||D-D_{pred}||_2. \tag{9} \label{eq9}
\end{equation}

As illustrated in Fig.~\ref{fig:depth}, it is evident that the depth priors significantly enhance the geometric structure, resulting in a more accurate depth map compared to both BAGS~\cite{peng2024bags} and the variant of our method without depth priors. In Section~\ref{sec:5.3}, we further conduct an ablation study comparing several other depth supervision strategies to demonstrate the effectiveness of our approach.

\begin{table*}[t]
\caption{
\textbf{Quantitative comparisons on the defocus blur dataset of Deblur NeRF~\cite{ma2022deblur}.} 
We report the PSNR, SSIM, and LPIPS metrics and color each cell as \colorbox{orange!50}{best} and \colorbox{yellow!50}{second best}. 
% BAGS$^\dag$ is reproduced from the official released code and BAGS is directly used from the results shown in the original paper. 
Our method outperforms other existing approaches across most scenes.
}
\centering
\resizebox{\textwidth}{!}{
\setlength{\tabcolsep}{0.8mm}{
\begin{tabular}{lcccccccccccccccccc}
\toprule
\multirow{2}{*}{Method} & \multicolumn{3}{c}{Deblur-NeRF~\cite{ma2022deblur}} & \multicolumn{3}{c}{DoF-NeRF~\cite{wu2022dof}} & \multicolumn{3}{c}{DP-NeRF~\cite{lee2023dp}} & \multicolumn{3}{c}{BAGS~\cite{peng2024bags}} & \multicolumn{3}{c}{Deblurring 3DGS~\cite{lee2024deblurring}}& \multicolumn{3}{c}{\textbf{Ours}}\\
& PSNR & SSIM & LPIPS & PSNR & SSIM & LPIPS & PSNR & SSIM & LPIPS & PSNR & SSIM & LPIPS & PSNR & SSIM & LPIPS& PSNR & SSIM & LPIPS\\
\midrule
Cake &26.27& 0.780& 0.128& 24.53& 0.731& 0.206 & 26.16& 0.778& 0.127& \cellcolor{orange!50}27.21 & \cellcolor{orange!50}0.818 & \cellcolor{yellow!50}0.108 & \cellcolor{yellow!50}26.88 & 0.803 &0.116 &26.83
&\cellcolor{yellow!50}0.808 &\cellcolor{orange!50}0.101\\  
Caps & 23.87& 0.713& 0.161& 22.67& 0.636& 0.261& 23.95& 0.712& \cellcolor{orange!50}0.143&24.16 &0.725 &0.159 &\cellcolor{yellow!50}24.50 &\cellcolor{yellow!50}0.742 & 0.149& \cellcolor{orange!50}24.62& \cellcolor{orange!50}0.749& \cellcolor{yellow!50}0.147\\ 
Cisco & 20.83& 0.727& 0.087& 20.64& 0.724& 0.107& 20.73& 0.726& 0.084&20.79 &\cellcolor{yellow!50}0.743 &\cellcolor{yellow!50}0.070 &\cellcolor{yellow!50}20.83 &0.732 & 0.079& \cellcolor{orange!50}21.00&\cellcolor{orange!50} 0.744& \cellcolor{orange!50}0.069\\
Coral & 19.85& 0.600& 0.121& 19.83& 0.570& 0.240& 20.11& 0.611& 0.118&\cellcolor{orange!50} 20.53& \cellcolor{yellow!50}0.628&\cellcolor{yellow!50}0.111& 19.78& 0.608& 0.131&\cellcolor{yellow!50} 20.37& \cellcolor{orange!50}0.630& \cellcolor{orange!50}0.109\\
Cupcake & 22.26& 0.722& 0.116& 21.89& 0.706& 0.143& 22.80& 0.741&  0.096& \cellcolor{yellow!50}22.93& \cellcolor{orange!50}0.762& \cellcolor{yellow!50}0.080& 22.11& 0.734& 0.099& \cellcolor{orange!50}22.97& \cellcolor{yellow!50}0.757& \cellcolor{orange!50}0.079\\
Cups & 26.21 &0.799 & 0.127& 25.26& 0.765& 0.202& \cellcolor{orange!50}26.75& 0.814& 0.104& 26.27&\cellcolor{yellow!50} 0.823& 0.104& \cellcolor{yellow!50}26.28&\cellcolor{orange!50}0.824& \cellcolor{yellow!50}0.103& 26.01& 0.817&\cellcolor{orange!50} 0.100\\
Daisy &23.52 &0.687 & 0.121& 23.22& 0.658& 0.194& \cellcolor{yellow!50}23.79& 0.697& 0.108& 23.74& \cellcolor{orange!50}0.746& \cellcolor{orange!50}0.062& 23.54& 0.731& 0.095& \cellcolor{orange!50}23.93& \cellcolor{yellow!50}0.735& \cellcolor{yellow!50}0.071\\
Sausage & 18.01 &0.500 & 0.180& 17.86& 0.488& 0.280& 18.35& 0.544& 0.147& 18.76& \cellcolor{yellow!50}0.574& \cellcolor{orange!50}0.110& \cellcolor{yellow!50}18.99 & 0.570& 0.141& \cellcolor{orange!50}19.11& \cellcolor{orange!50}0.576& \cellcolor{yellow!50}0.119\\
Seal & 26.04& 0.777& 0.105& 24.85& 0.687& 0.143& 25.95& 0.778& 0.103&\cellcolor{yellow!50} 26.52&\cellcolor{yellow!50} 0.812&\cellcolor{yellow!50} 0.090& 26.18& 0.817& 0.098&\cellcolor{orange!50} 26.57& \cellcolor{orange!50}0.825& \cellcolor{orange!50}0.087\\
Tools & 27.81& 0.895& 0.061& 26.21& 0.854& 0.128& 28.07& 0.898& 0.054&\cellcolor{orange!50}28.60 & \cellcolor{yellow!50}0.913& \cellcolor{orange!50}0.046& 27.96&\cellcolor{yellow!50} 0.910& 0.058&\cellcolor{yellow!50} 28.29& \cellcolor{orange!50}0.913& \cellcolor{yellow!50}0.051\\
% \hline
% Average& 23.47& 0.720& 0.121& & & & & & & \cellcolor{yellow!50}23.95& \cellcolor{yellow!50}0.754& \cellcolor{yellow!50}0.095& 23.71& 0.747& 0.107& \cellcolor{orange!50}23.96& \cellcolor{orange!50}0.755& \cellcolor{orange!50}0.093\\
\bottomrule
\end{tabular} }
}
\label{tab:quantitative}
\end{table*}

\subsection{Defocus-to-Focus Adaptation}
\label{sec:4.3}
% \begin{figure}[!t]
%   \centering
%   \setlength{\belowcaptionskip}{-5pt}
%   \includegraphics[width=\linewidth]{figs/adaptation.pdf}
%   \caption{\normalsize{\textbf{The process of defocus-to-focus adaptation.} According to the depth map $D$, focus distance $\mathcal{F}$, and reweighting function $\Psi$, we derive the weight map $\Psi(D)$ to adjust the aperture size and loss function.}
%   }
%   \label{fig:adaptation}
% \end{figure}
The CoC effect achieved under ideal optical imaging conditions, as described in Algorithm~\ref{alg1}, inevitably differs from the real CoC effect produced by an actual DSLR camera. This discrepancy hinders our ability to accurately model the out-of-focus blur in shallow DoF input images.

To achieve sharper scenes, we propose an adaptation process that allows transitions from defocus to focus. 
Specifically, modeling the defocus effects with bokeh blur facilitates learning accurate lens parameters.
Consequently, we place greater emphasis on capturing the sharp areas in training images once the lens parameters $\mathcal{A}$ and $\mathcal{F}$ have been optimized and converged. This approach is effective because, when the focus distance is determined, the sharp regions in the scene become approximately identifiable.
After $t$ optimization iterations, we reweight the $\mathcal{L}_{rec}$ and $\mathcal{L}_{depth}$ loss by employing a step-like function:
\begin{equation}
\Psi(x) =\left\{
\begin{array}{cl}
1 &, \mathrm{iterations}<t\\
1/(1+e^{-a\cdot(x-b)}) &, \mathrm{iterations} \geq t\\
\end{array}
\right.
      \tag{10} \label{eq10}
\end{equation}
where $a$ and $b$ are hyperparameters, and $x = |\frac{1}{\mathcal{F}}-\frac{1}{d}|$.

In addition, the aperture size $\mathcal{A}$ is generally large after the convergence, which can cause blurring of pixels that should be in focus. Hence, we also reweight the aperture size $\mathcal{A}$ on a pixel-wise basis: 
\begin{equation}
    \mathcal{A}' = \mathcal{A} \cdot \Psi. \tag{11} \label{eq11}
\end{equation}

\noindent \textbf{Full Objective.} We derive the final loss terms by incorporating the reconstruction loss from Eqn.~\ref{eq7}, the depth loss from Eqn.~\ref{eq9}, and an additional normal consistency loss as proposed by Huang~\etal~\cite{huang20242d}. The normal consistency loss ensures that the Gaussian splats align with the surface by measuring the consistency between the normal directions computed from the Gaussian and the depth map:
\begin{equation}
    \mathcal{L}_{normal}=\sum_iw_i(1-n_i^{\mathsf{T}}\hat{n}_i), \tag{12} \label{eq12}
\end{equation}
where $\hat{n}$ represents the surface normal direction obtained by applying finite differences on the depth map, $i$ indexes the intersected splats along the ray and $w$ denotes the blending weight of the intersection point. 

Our final training loss $\mathcal{L}$ is:
\begin{equation}
    \mathcal{L}=\Psi\odot(\mathcal{L}_{rec}+w_d\mathcal{L}_{depth})+w_n\mathcal{L}_{normal}, \tag{13} \label{eq13}
\end{equation}
where $\odot$ is Hadamard product.
\section{Experiments}
\label{sec:exp}
%\subsection{Settings}
\noindent \textbf{Implementation Details.}
In our experiments, we set the smoothness of confuse transition $\alpha=4$. In addition, we set $a = 15$, $b = 0.3$, and $t=10000$ for the defocus-to-focus adaptation. 
Our method is built upon Mip-Splatting~\cite{yu2024mip}, and our optimization strategy and hyperparameter settings remain consistent with it. 
We train each scene for $30000$ iterations and set loss weights $w_d=0.01$ and $w_n=0.05$.

\noindent \textbf{Baselines and Evaluation Metrics.} We compare our method with state-of-the-art deblurring techniques~\cite{peng2024bags,ma2022deblur,lee2024deblurring} and a depth-of-field method~\cite{wu2022dof}. To quantitatively evaluate the quality of novel view images, we adopt widely used metrics such as PSNR, SSIM~\cite{wang2004image}, and LPIPS~\cite{zhang2018unreasonable}. We will compare with Wang~\etal~\cite{wang2024dof} once they release the code or provide the full experimental results in their paper.

\noindent \textbf{Datasets.}
Following~\cite{peng2024bags,lee2024deblurring,ma2022deblur}, we evaluate our model on the Deblur-NeRF dataset~\cite{ma2022deblur}.
Subsequently, we conduct further evaluations on the Real Forward-facing dataset~\cite{mildenhall2021nerf} and T\&T\_DB dataset~\cite{kerbl20233d}. The effectiveness of our method is validated on these datasets for both shallow DoF inputs and normal inputs. Furthermore, we also introduce a dataset in Section~\ref{sec:dataset} to evaluate our model more comprehensively.
\begin{table}[t]
\centering
\caption{
\textbf{Quantitative comparisons on our synthetic dataset.}
}
\resizebox{\columnwidth}{!}{
\begin{tabular}{lccccc}
\toprule
Method & PSNR$\uparrow$ & SSIM$\uparrow$ & LPIPS$\downarrow$ & $\delta_{\mathcal{A}}\downarrow$ & $\delta_{\mathcal{F}}\downarrow$\\
\midrule
% BAGS~\cite{peng2024bags}& 23.91& 0.704& 0.293&--- &---\\
DoF-NeRF~\cite{wu2022dof}& 25.59& 0.788& 0.207& 0.196& 0.256\\
Ours& \textbf{28.70}& \textbf{0.864}& \textbf{0.095}& \textbf{0.126}& \textbf{0.079}\\
\bottomrule
\end{tabular}

}
\vspace{-0.05in}
\label{tab:synthetic}
\end{table}
\subsection{A Synthetic Dataset}
\label{sec:dataset}
\begin{figure}[!t]
  \centering
  \setlength{\belowcaptionskip}{-5pt}
  \includegraphics[width=\linewidth]{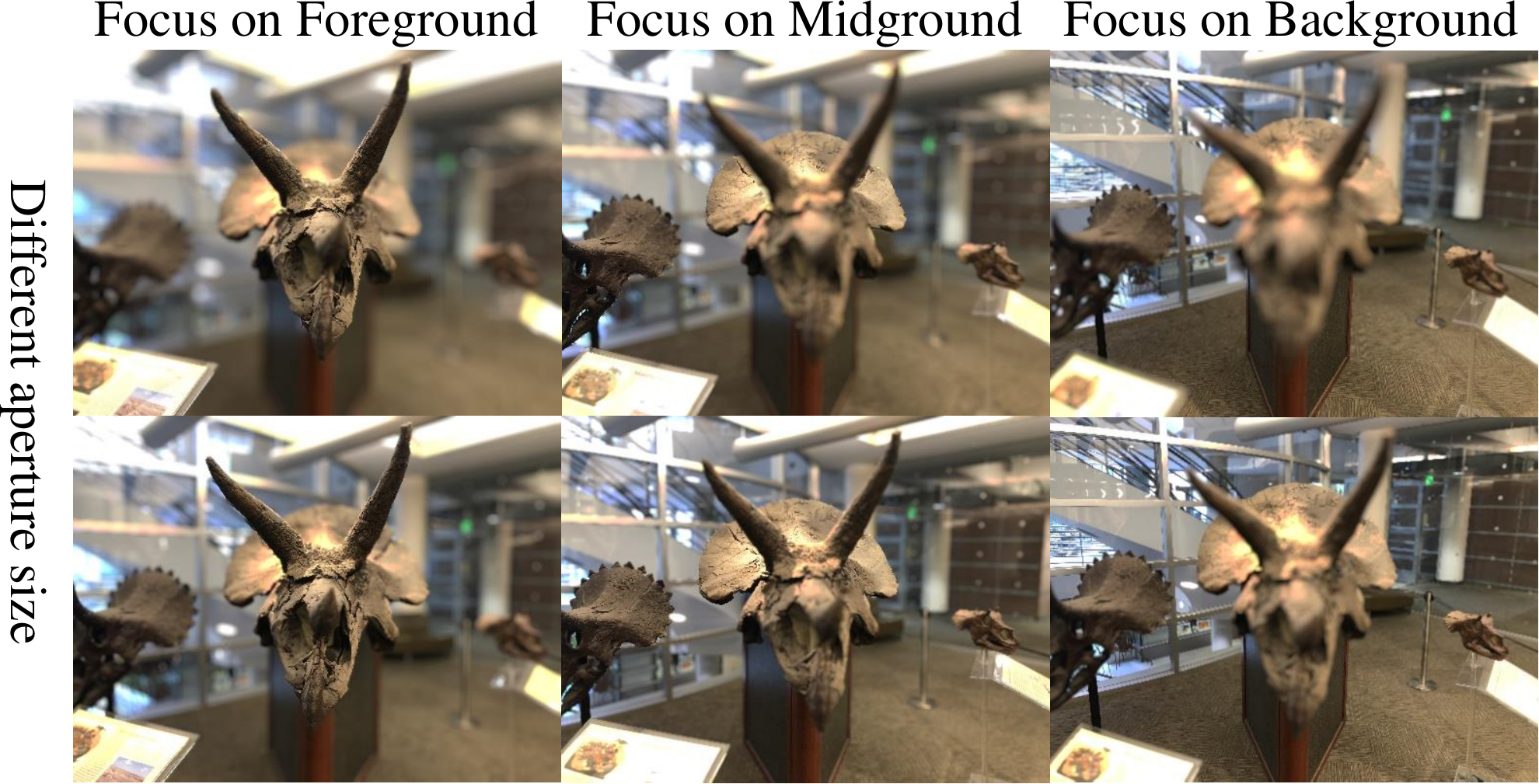}
  \caption{\textbf{Examples in our dataset.} For each scene, we apply a DoF rendering method~\cite{peng2022bokehme} to convert all-in-focus images to shallow DoF images with varying focus locations and aperture sizes. This dataset is used to evaluate the model's refocusing ability.
  }
  \label{fig:synthetic}
\end{figure}

We introduce a synthetic dataset, as illustrated in Fig.~\ref{fig:synthetic}, based on depth estimation~\cite{bochkovskii2024depth} and a single-image DoF rendering method~\cite{peng2022bokehme} for each image in the Real Forward Facing dataset~\cite{mildenhall2021nerf} and T\&T\_DB dataset~\cite{kerbl20233d}. This allows us to evaluate the refocusing ability of our model and to determine whether it accurately learns the correct aperture size and focus distance.
In particular, by feeding the DoF rendering method~\cite{peng2022bokehme} the known aperture size and focus distance, we can convert the all-in-focus images into shallow DoF images. Unlike existing defocus blur datasets proposed by Ma \etal~\cite{ma2022deblur} and Wu \etal~\cite{wu2022dof}, the test set in our dataset comprises images with shallow DoF. With known lens parameters, our model fits this shallow DoF effect to quantitatively measure the refocusing ability. Meanwhile, we establish the known lens parameters beforehand as ground truth, allowing us to calculate the errors $\delta_\mathcal{A}$ and $\delta_\mathcal{F}$ between the learned parameters and the ground truth.
More details can be found in the supplementary material.
% We hope that this can bring controllable depth-of-field into the sight of a broader 3D Gaussian Splatting community and motivate further research.

\subsection{Comparisons}
\begin{figure*}[!t]
  \centering
  \setlength{\belowcaptionskip}{-5pt}
  \includegraphics[width=\linewidth]{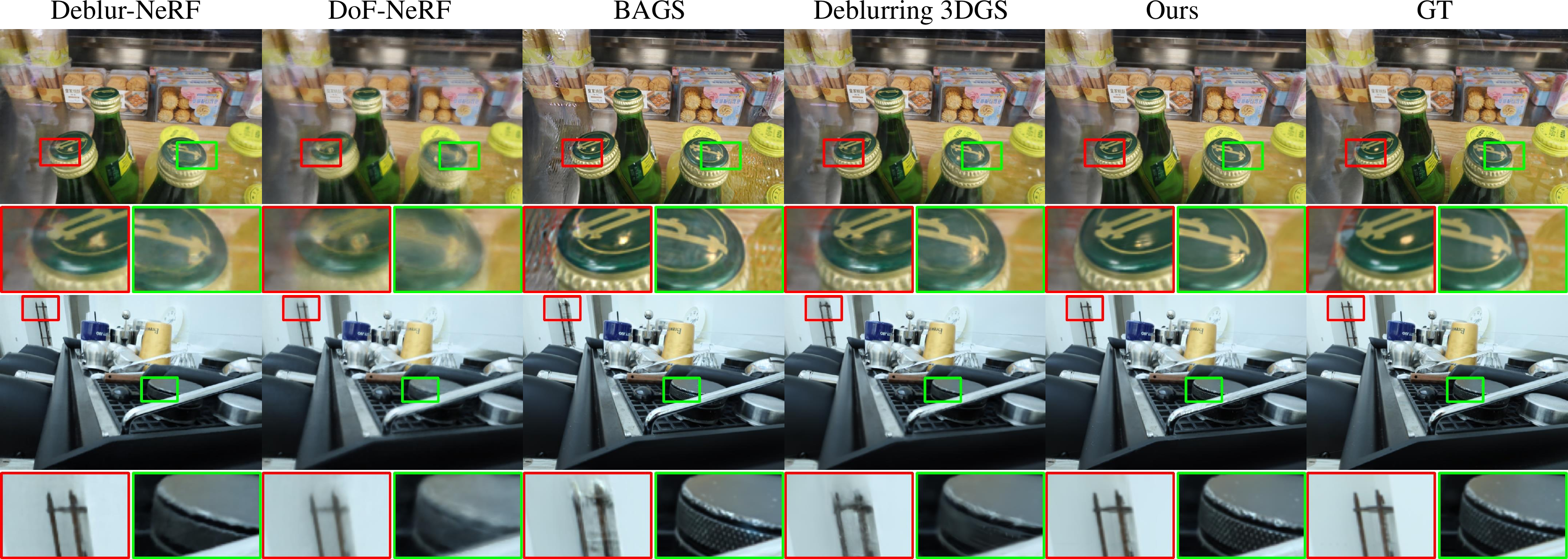}
  \caption{\textbf{Qualitative comparisons against all baselines.} Compared to other state-of-the-art methods, our method represents sharper scenes and generates novel view images with less blur.
  }
  \label{fig:qualitative}
\end{figure*}
\noindent \textbf{Quantitative Comparisons.}
We quantitatively compare \textit{DoF-Gaussian} against other state-of-the-art methods using a real-world defocus blur dataset~\cite{ma2022deblur}. As shown in Table~\ref{tab:quantitative}, our method outperforms other state-of-the-art baselines. These results suggest that our method achieves superior defocus deblurring performance and produces higher-quality novel view synthesis images.

\noindent \textbf{Qualitative Comparisons.}
Visual qualitative comparisons are presented in Fig.~\ref{fig:qualitative}. Our method surpasses all other methods in terms of image fidelity, generating novel view images that are more faithful to the ground truth images and exhibit less blur. Specifically, the bottle caps and shelves are the sharpest in our rendered images.

\noindent \textbf{All-in-focus Inputs.} 
\begin{table}[t]
\centering
\caption{
\textbf{Comparisons in all-in-focus settings.} We compare our method with Mip-Splatting on all-in-focus datasets to validate the effectiveness of our model under general input conditions.}

\resizebox{0.9\columnwidth}{!}{

\begin{tabular}{lccc}
\toprule
{Method} & PSNR$\uparrow$ & SSIM$\uparrow$ & LPIPS$\downarrow$ \\
\midrule
Mip-Splatting~\cite{yu2024mip}& 27.05& 0.893& \textbf{0.115}\\
Ours&\textbf{27.81}& \textbf{0.902}& 0.117\\
\bottomrule
\end{tabular}
}
\vspace{-0.05in}
\label{tab:all-in-focus}
\end{table}
To further validate the effectiveness of our method in the all-in-focus setting, we design an experiment using wide DoF images as inputs. We conduct a comparison between our method and Mip-Splatting~\cite{yu2024mip} on two all-in-focus datasets, the Real Forward Facing dataset~\cite{mildenhall2021nerf} and the T\&T\_DB dataset~\cite{kerbl20233d}. As shown in Table~\ref{tab:all-in-focus}, our method achieves comparable or even better performance to Mip-Splatting on the average metrics across the datasets. This demonstrates that our lens-based model not only excels with shallow DoF inputs but also achieves good results with general input images.

\noindent \textbf{Our Synthetic Dataset.} 
As shown in Table~\ref{tab:synthetic}, our method achieves superior refocusing ability through controlled DoF rendering, producing rendered images with the highest quality on our synthetic dataset. Furthermore, our method learns more precise aperture size and focus distance than DoF-NeRF~\cite{wu2022dof}. Since Deblur-NeRF~\cite{ma2022deblur}, BAGS~\cite{peng2024bags}, and Deblurring 3DGS~\cite{lee2024deblurring} do not incorporate a lens model, they cannot generate novel view images with a specific DoF. As a result, these methods cannot be evaluated on this dataset due to their lack of refocusing capability.

\begin{figure*}[!t]
  \centering
  \setlength{\belowcaptionskip}{-5pt}
  \includegraphics[width=\linewidth]{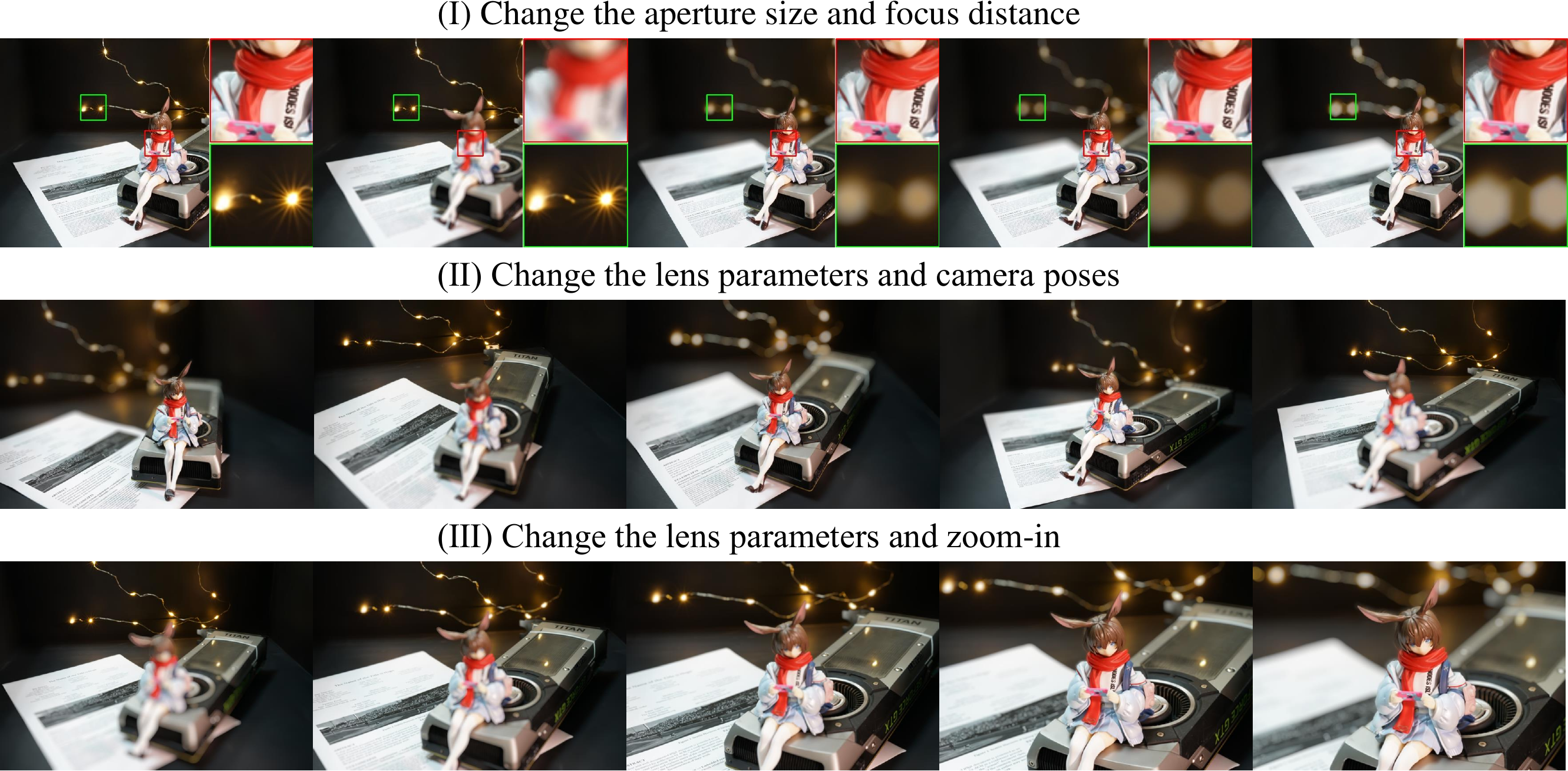}
  \caption{\textbf{Applications of our controllable DoF effects.} Users can create their own cinematic moments by combining changes in aperture size, focus distance, camera poses, and zoom.
  }\vspace{-5pt}
  \label{fig:applications}
\end{figure*}

\subsection{Ablation Study}
\label{sec:5.3}
\begin{figure}[htb]
  \centering
  \setlength{\belowcaptionskip}{-5pt}
  \includegraphics[width=\linewidth]{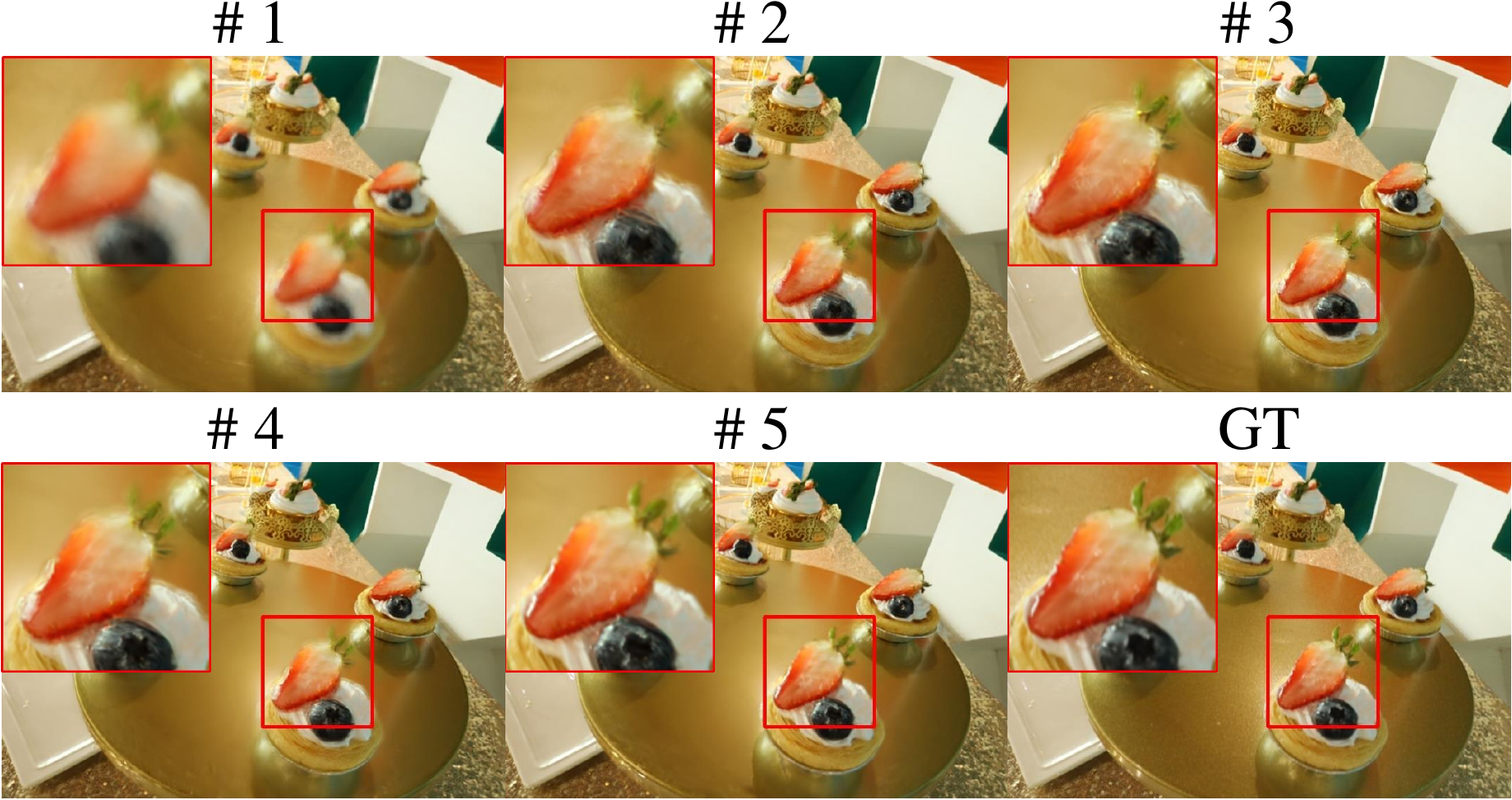}
  \caption{\textbf{The visualizations of the ablation study.} Qualitative comparisons show that the full model ($\#5$) yields the best result and the reconstructed strawberry is the sharpest.
  }
  \label{fig:ablation}
\end{figure}
As shown in Table~\ref{tab:ablation} and Fig.~\ref{fig:ablation}, we conduct ablation studies to evaluate the effectiveness of our designs. The comparison between $\#1$ and $\#2$ indicates that the lens-based imaging model has a significant impact. This lens model not only facilitates defocus deblurring but also provides the ability to control the depth of field. Next, the defocus-to-focus adaptation ($\#3$) and the per-scene adjustment of depth priors ($\#4$) further enhance the scene geometry and improve the deblurring of details. Ultimately, the combination of all components yields the highest performance gain, leading to a $2.66$ dB increase in PSNR over the baseline model.

In addition, we conduct an ablation study on different depth supervision methods to demonstrate the superiority of our per-scene adjustment of depth priors, as shown in Table~\ref{tab:ablation_depth}. Direct supervision using depth maps predicted by the depth network without scene-specific fine-tuning hinders the reconstruction quality of 3D-GS. Likewise, supervision based on sparse depth maps projected from a 3D point cloud fails to yield optimal results.
We show visualizations of depth maps rendered by different depth supervision strategies in the supplementary material.
% \begin{table}[t]
% \centering
% \resizebox{\columnwidth}{!}{

% \begin{tabular}{lccc}
% \toprule
% Method & PSNR$\uparrow$ & SSIM$\uparrow$ & LPIPS$\downarrow$ \\
% \midrule
% baseline& 21.31& 0.636& 0.239\\
% w/o lens-based model& & & \\
% w/o depth supervision& 23.59& 0.742& 0.104\\
% w/o defocus-to-focus adaptation& 23.42& 0.738& 0.098\\
% Full model & \textbf{23.97}& \textbf{0.756}& \textbf{0.093}\\
% \bottomrule
% \end{tabular}

% }
% \caption{
% Ablation study on each component of our method.
% }
% \vspace{-0.05in}
% \label{tab:template}
% \end{table}
\begin{table}[t]
\centering
\caption{
\textbf{Ablation study on each component of our method.}
}
\resizebox{\columnwidth}{!}{

\begin{tabular}{l|ccc|ccc}
\toprule
 &Lens & Depth & Defocus-to-focus&PSNR$\uparrow$ & SSIM$\uparrow$ & LPIPS$\downarrow$ \\
\midrule
$\#1$ &\XSolidBrush& \XSolidBrush& \XSolidBrush& 21.31& 0.636& 0.239\\
$\#2$ &\Checkmark & \XSolidBrush& \XSolidBrush& 23.05& 0.728& 0.109\\
$\#3$ & \Checkmark& \XSolidBrush& \Checkmark& 23.59& 0.742& 0.104\\
$\#4$& \Checkmark & \Checkmark& \XSolidBrush& 23.42& 0.738& 0.098\\
$\#5$& \Checkmark& \Checkmark& \Checkmark& \textbf{23.97}& \textbf{0.756}& \textbf{0.093}\\
\bottomrule
\end{tabular}

}
\vspace{-0.05in}
\label{tab:ablation}
\end{table}
\begin{table}[t]
\centering
\caption{
\textbf{Ablation study on different depth priors.}
}
\resizebox{0.9\columnwidth}{!}{

\begin{tabular}{lccc}
\toprule
{Method} & PSNR$\uparrow$ & SSIM$\uparrow$ & LPIPS$\downarrow$ \\
\midrule
No fine-tuned depth& 23.53& 0.739& 0.114\\
Sparse depth& 23.44& 0.737& 0.113\\
Ours&\textbf{23.97}& \textbf{0.756}& \textbf{0.093}\\
\bottomrule
\end{tabular}
}
\vspace{-0.05in}
\label{tab:ablation_depth}
\end{table}
\subsection{Interactive Applications}
Thanks to the lens-based model, we can not only represent 3D scenes from shallow DoF inputs but also render interactive DoF effects (see Fig.~\ref{fig:applications}). 
Various cinematography styles can be achieved through different camera configurations, and users can input custom datasets, whether all-in-focus or bokeh images.
By controlling aperture size and focus distance, we can render novel view images with different depths of field. In addition, we can modify the shape of the CoC, such as changing it from a circle to a hexagon.
Users can also adjust the depth of field while moving the camera or zooming to create cinematic effects. Unlike~\cite{wu2022dof}, our GS framework significantly enhances the training and rendering efficiency, allowing for faster and more seamless interactions. The details on processing time will be discussed in the supplementary material.

\section{Conclusion}
\label{sec:conclusion}

While 3D Gaussian Splatting methods have achieved impressive performance on a wide range of reconstruction tasks, the applications on controllable DoF effects remain challenging and understudied. In this paper, we propose DoF-Gaussian, a controllable DoF method for 3D Gaussian Splatting. Specifically, we develop a lens-based model rather than pinhole imaging to overcome the limitations imposed by shallow DoF inputs for 3D-GS. Furthermore, we propose the per-scene depth adjustment and a defocus-to-focus adaptation to guarantee the performance of defocus deblurring. We also introduce a synthetic dataset for a more comprehensive evaluation. Thanks to the imaging principles, our method supports various interactive applications.

\noindent \textbf{Future work.} Inspired by our experiments, we found that modeling real-world physical imaging principles can enable our method not only to handle shallow DoF inputs but also to perform effectively on general inputs.
This insight motivates us to explore a combination of non-ideal conditions, such as shallow DoF inputs, sparse views, and varied lighting environments, which are more aligned with casual photography in our daily lives. 
% We hope to bring controllable DoF into the sight of a broader community in novel view synthesis and motivate further research. \quad

\noindent \textbf{Acknowledgement.}
This study is supported under the RIE2020 Industry Alignment Fund – Industry Collaboration Projects (IAF-ICP) Funding Initiative, as well as cash and in-kind contribution from the industry partner(s).

{
    \small
    \bibliographystyle{ieeenat_fullname}
    \bibliography{main}

\begin{thebibliography}{62}
\providecommand{\natexlab}[1]{#1}
\providecommand{\url}[1]{\texttt{#1}}
\expandafter\ifx\csname urlstyle\endcsname\relax
  \providecommand{\doi}[1]{doi: #1}\else
  \providecommand{\doi}{doi: \begingroup \urlstyle{rm}\Url}\fi

\bibitem[Abadie et~al.(2018)Abadie, McAuley, Golubev, Hill, and Lagarde]{abadie2018advances}
Guillaume Abadie, Steve McAuley, Evegenii Golubev, Stephen Hill, and Sebastien Lagarde.
\newblock {Advances in real-time rendering in games}.
\newblock In \emph{ACM SIGGRAPH 2018 Courses}. 2018.

\bibitem[Bochkovskii et~al.(2024)Bochkovskii, Delaunoy, Germain, Santos, Zhou, Richter, and Koltun]{bochkovskii2024depth}
Aleksei Bochkovskii, Ama{\"e}l Delaunoy, Hugo Germain, Marcel Santos, Yichao Zhou, Stephan~R Richter, and Vladlen Koltun.
\newblock Depth pro: Sharp monocular metric depth in less than a second.
\newblock \emph{arXiv preprint arXiv:2410.02073}, 2024.

\bibitem[Busam et~al.(2019)Busam, Hog, McDonagh, and Slabaugh]{busam2019sterefo}
Benjamin Busam, Matthieu Hog, Steven McDonagh, and Gregory Slabaugh.
\newblock {Sterefo: Efficient image refocusing with stereo vision}.
\newblock In \emph{ICCVW}, 2019.

\bibitem[Chen and Liu(2024)]{chen2024deblur}
Wenbo Chen and Ligang Liu.
\newblock {Deblur-GS: 3D Gaussian Splatting from Camera Motion Blurred Images}.
\newblock \emph{PACMCGIT}, 7\penalty0 (1), 2024.

\bibitem[Chung et~al.(2024)Chung, Oh, and Lee]{chung2024depth}
Jaeyoung Chung, Jeongtaek Oh, and Kyoung~Mu Lee.
\newblock {Depth-regularized optimization for 3D Gaussian Splatting in few-shot images}.
\newblock In \emph{CVPR}, 2024.

\bibitem[Conde et~al.(2023)Conde, Kolmet, Seizinger, Bishop, Timofte, Kong, Zhang, Wu, Wang, Peng, et~al.]{conde2023lens}
Marcos~V Conde, Manuel Kolmet, Tim Seizinger, Tom~E Bishop, Radu Timofte, Xiangyu Kong, Dafeng Zhang, Jinlong Wu, Fan Wang, Juewen Peng, et~al.
\newblock {Lens-to-lens bokeh effect transformation. NTIRE 2023 challenge report}.
\newblock In \emph{CVPR}, 2023.

\bibitem[Darmon et~al.(2024)Darmon, Porzi, Rota-Bul{\`o}, and Kontschieder]{darmon2024robust}
Fran{\c{c}}ois Darmon, Lorenzo Porzi, Samuel Rota-Bul{\`o}, and Peter Kontschieder.
\newblock Robust gaussian splatting.
\newblock \emph{arXiv preprint arXiv:2404.04211}, 2024.

\bibitem[Dutta et~al.(2021)Dutta, Das, Shah, and Tiwari]{dutta2021stacked}
Saikat Dutta, Sourya~Dipta Das, Nisarg~A Shah, and Anil~Kumar Tiwari.
\newblock {Stacked deep multi-scale hierarchical network for fast bokeh effect rendering from a single image}.
\newblock In \emph{CVPR}, 2021.

\bibitem[Eigen et~al.(2014)Eigen, Puhrsch, and Fergus]{eigen2014depth}
David Eigen, Christian Puhrsch, and Rob Fergus.
\newblock {Depth map prediction from a single image using a multi-scale deep network}.
\newblock \emph{NeurIPS}, 27, 2014.

\bibitem[Hecht(2012)]{hecht2012optics}
Eugene Hecht.
\newblock \emph{{Optics}}.
\newblock Pearson Education India, 2012.

\bibitem[Huang et~al.(2024)Huang, Yu, Chen, Geiger, and Gao]{huang20242d}
Binbin Huang, Zehao Yu, Anpei Chen, Andreas Geiger, and Shenghua Gao.
\newblock {2D Gaussian Splatting for geometrically accurate radiance fields}.
\newblock In \emph{ACM SIGGRAPH 2024 Conference Papers}, 2024.

\bibitem[Ignatov et~al.(2019)Ignatov, Patel, Timofte, Zheng, Ye, Huang, Tian, Dutta, Purohit, Kandula, et~al.]{ignatov2019aim}
Andrey Ignatov, Jagruti Patel, Radu Timofte, Bolun Zheng, Xin Ye, Li Huang, Xiang Tian, Saikat Dutta, Kuldeep Purohit, Praveen Kandula, et~al.
\newblock {Aim 2019 challenge on bokeh effect synthesis: Methods and results}.
\newblock In \emph{ICCVW}. IEEE, 2019.

\bibitem[Ignatov et~al.(2020{\natexlab{a}})Ignatov, Patel, and Timofte]{ignatov2020rendering}
Andrey Ignatov, Jagruti Patel, and Radu Timofte.
\newblock {Rendering natural camera bokeh effect with deep learning}.
\newblock In \emph{CVPRW}, 2020{\natexlab{a}}.

\bibitem[Ignatov et~al.(2020{\natexlab{b}})Ignatov, Timofte, Qian, Qiao, Lin, Guo, Li, Leng, Cheng, Peng, et~al.]{ignatov2020aim}
Andrey Ignatov, Radu Timofte, Ming Qian, Congyu Qiao, Jiamin Lin, Zhenyu Guo, Chenghua Li, Cong Leng, Jian Cheng, Juewen Peng, et~al.
\newblock {AIM 2020 challenge on rendering realistic bokeh}.
\newblock In \emph{ECCVW}. Springer, 2020{\natexlab{b}}.

\bibitem[Jin et~al.(2024)Jin, Jiao, Duan, Yang, Guo, Ren, and Li]{jin2024lighting}
Xin Jin, Pengyi Jiao, Zheng-Peng Duan, Xingchao Yang, Chun-Le Guo, Bo Ren, and Chongyi Li.
\newblock {Lighting Every Darkness with 3DGS: Fast Training and Real-Time Rendering for HDR View Synthesis}.
\newblock \emph{arXiv preprint arXiv:2406.06216}, 2024.

\bibitem[Kerbl et~al.(2023)Kerbl, Kopanas, Leimk{\"u}hler, and Drettakis]{kerbl20233d}
Bernhard Kerbl, Georgios Kopanas, Thomas Leimk{\"u}hler, and George Drettakis.
\newblock {3D Gaussian Splatting for Real-Time Radiance Field Rendering.}
\newblock \emph{ACM TOG}, 42\penalty0 (4), 2023.

\bibitem[Kim et~al.(2024)Kim, Gu, and Choo]{kim2024lensnerf}
Min-Jung Kim, Gyojung Gu, and Jaegul Choo.
\newblock Lens{NeRF}: Rethinking volume rendering based on thin-lens camera model.
\newblock In \emph{WACV}, 2024.

\bibitem[Lee et~al.(2024)Lee, Lee, Sun, Ali, and Park]{lee2024deblurring}
Byeonghyeon Lee, Howoong Lee, Xiangyu Sun, Usman Ali, and Eunbyung Park.
\newblock {Deblurring 3D Gaussian Splatting}.
\newblock \emph{arXiv preprint arXiv:2401.00834}, 2024.

\bibitem[Lee et~al.(2023)Lee, Lee, Shin, and Lee]{lee2023dp}
Dogyoon Lee, Minhyeok Lee, Chajin Shin, and Sangyoun Lee.
\newblock {Dp-NeRF: Deblurred neural radiance field with physical scene priors}.
\newblock In \emph{CVPR}, 2023.

\bibitem[Lee et~al.(2010)Lee, Eisemann, and Seidel]{lee2010real}
Sungkil Lee, Elmar Eisemann, and Hans-Peter Seidel.
\newblock {Real-time lens blur effects and focus control}.
\newblock \emph{ACM TOG}, 29\penalty0 (4), 2010.

\bibitem[Li et~al.(2024)Li, Zhang, Bai, Zheng, Ning, Zhou, and Gu]{li2024dngaussian}
Jiahe Li, Jiawei Zhang, Xiao Bai, Jin Zheng, Xin Ning, Jun Zhou, and Lin Gu.
\newblock {DNGaussian: Optimizing sparse-view 3D Gaussian radiance fields with global-local depth normalization}.
\newblock In \emph{CVPR}, 2024.

\bibitem[Li et~al.(2019)Li, Dekel, Cole, Tucker, Snavely, Liu, and Freeman]{li2019learning}
Zhengqi Li, Tali Dekel, Forrester Cole, Richard Tucker, Noah Snavely, Ce Liu, and William~T Freeman.
\newblock {Learning the depths of moving people by watching frozen people}.
\newblock In \emph{CVPR}, 2019.

\bibitem[Liu et~al.(2025)Liu, Wang, Hu, Shen, Ye, Zang, Cao, Li, and Liu]{liu2025mvsgaussian}
Tianqi Liu, Guangcong Wang, Shoukang Hu, Liao Shen, Xinyi Ye, Yuhang Zang, Zhiguo Cao, Wei Li, and Ziwei Liu.
\newblock {MVSGaussian: Fast Generalizable Gaussian Splatting Reconstruction from Multi-View Stereo}.
\newblock In \emph{ECCV}. Springer, 2025.

\bibitem[Luo et~al.(2024{\natexlab{a}})Luo, Sun, Peng, and Cao]{luo2024dynamic}
Xianrui Luo, Huiqiang Sun, Juewen Peng, and Zhiguo Cao.
\newblock {Dynamic Neural Radiance Field From Defocused Monocular Video}.
\newblock \emph{arXiv preprint arXiv:2407.05586}, 2024{\natexlab{a}}.

\bibitem[Luo et~al.(2024{\natexlab{b}})Luo, Shi, Shen, Huang, Ye, Peng, and Cao]{luo2024video}
Yawen Luo, Min Shi, Liao Shen, Yachuan Huang, Zixuan Ye, Juewen Peng, and Zhiguo Cao.
\newblock Video bokeh rendering: Make casual videography cinematic.
\newblock In \emph{ACM MM}, 2024{\natexlab{b}}.

\bibitem[Ma et~al.(2022)Ma, Li, Liao, Zhang, Wang, Wang, and Sander]{ma2022deblur}
Li Ma, Xiaoyu Li, Jing Liao, Qi Zhang, Xuan Wang, Jue Wang, and Pedro~V Sander.
\newblock Deblur-{NeRF}: Neural radiance fields from blurry images.
\newblock In \emph{CVPR}, 2022.

\bibitem[Martin-Brualla et~al.(2021)Martin-Brualla, Radwan, Sajjadi, Barron, Dosovitskiy, and Duckworth]{martin2021nerf}
Ricardo Martin-Brualla, Noha Radwan, Mehdi~SM Sajjadi, Jonathan~T Barron, Alexey Dosovitskiy, and Daniel Duckworth.
\newblock {NeRF in the wild: Neural radiance fields for unconstrained photo collections}.
\newblock In \emph{CVPR}, 2021.

\bibitem[Mildenhall et~al.(2021)Mildenhall, Srinivasan, Tancik, Barron, Ramamoorthi, and Ng]{mildenhall2021nerf}
Ben Mildenhall, Pratul~P Srinivasan, Matthew Tancik, Jonathan~T Barron, Ravi Ramamoorthi, and Ren Ng.
\newblock {NeRF: Representing scenes as neural radiance fields for view synthesis}.
\newblock \emph{Communications of the ACM}, 65\penalty0 (1), 2021.

\bibitem[Mildenhall et~al.(2022)Mildenhall, Hedman, Martin-Brualla, Srinivasan, and Barron]{mildenhall2022nerf}
Ben Mildenhall, Peter Hedman, Ricardo Martin-Brualla, Pratul~P Srinivasan, and Jonathan~T Barron.
\newblock {NeRF in the dark: High dynamic range view synthesis from noisy raw images}.
\newblock In \emph{CVPR}, 2022.

\bibitem[Moenne-Loccoz et~al.(2024)Moenne-Loccoz, Mirzaei, Perel, de~Lutio, Martinez~Esturo, State, Fidler, Sharp, and Gojcic]{moenne20243d}
Nicolas Moenne-Loccoz, Ashkan Mirzaei, Or Perel, Riccardo de Lutio, Janick Martinez~Esturo, Gavriel State, Sanja Fidler, Nicholas Sharp, and Zan Gojcic.
\newblock 3d gaussian ray tracing: Fast tracing of particle scenes.
\newblock \emph{ACM Transactions on Graphics (TOG)}, 43\penalty0 (6):\penalty0 1--19, 2024.

\bibitem[Park et~al.(2021)Park, Sinha, Barron, Bouaziz, Goldman, Seitz, and Martin-Brualla]{park2021nerfies}
Keunhong Park, Utkarsh Sinha, Jonathan~T Barron, Sofien Bouaziz, Dan~B Goldman, Steven~M Seitz, and Ricardo Martin-Brualla.
\newblock {Nerfies: Deformable neural radiance fields}.
\newblock In \emph{ICCV}, 2021.

\bibitem[Pedrotti et~al.(2018)Pedrotti, Pedrotti, and Pedrotti]{pedrotti2018introduction}
Frank~L Pedrotti, Leno~M Pedrotti, and Leno~S Pedrotti.
\newblock \emph{{Introduction to optics}}.
\newblock Cambridge University Press, 2018.

\bibitem[Peng and Chellappa(2023)]{peng2023pdrf}
Cheng Peng and Rama Chellappa.
\newblock {PDRF: progressively deblurring radiance field for fast scene reconstruction from blurry images}.
\newblock In \emph{AAAI}, 2023.

\bibitem[Peng et~al.(2024)Peng, Tang, Zhou, Wang, Liu, Li, and Chellappa]{peng2024bags}
Cheng Peng, Yutao Tang, Yifan Zhou, Nengyu Wang, Xijun Liu, Deming Li, and Rama Chellappa.
\newblock {BAGS: Blur Agnostic Gaussian Splatting through Multi-Scale Kernel Modeling}.
\newblock \emph{arXiv preprint arXiv:2403.04926}, 2024.

\bibitem[Peng et~al.(2022)Peng, Cao, Luo, Lu, Xian, and Zhang]{peng2022bokehme}
Juewen Peng, Zhiguo Cao, Xianrui Luo, Hao Lu, Ke Xian, and Jianming Zhang.
\newblock {Bokehme: When neural rendering meets classical rendering}.
\newblock In \emph{CVPR}, 2022.

\bibitem[Peng et~al.(2023)Peng, Pan, Liu, Luo, Sun, Shen, Xian, and Cao]{peng2023selective}
Juewen Peng, Zhiyu Pan, Chengxin Liu, Xianrui Luo, Huiqiang Sun, Liao Shen, Ke Xian, and Zhiguo Cao.
\newblock Selective bokeh effect transformation.
\newblock In \emph{Proceedings of the IEEE/CVF conference on computer vision and pattern recognition}, pages 1456--1464, 2023.

\bibitem[Pich{\'e}-Meunier et~al.(2023)Pich{\'e}-Meunier, Hold-Geoffroy, Zhang, and Lalonde]{piche2023lens}
Dominique Pich{\'e}-Meunier, Yannick Hold-Geoffroy, Jianming Zhang, and Jean-Fran{\c{c}}ois Lalonde.
\newblock Lens parameter estimation for realistic depth of field modeling.
\newblock In \emph{Proceedings of the IEEE/CVF International Conference on Computer Vision}, pages 499--508, 2023.

\bibitem[Qian et~al.(2020)Qian, Qiao, Lin, Guo, Li, Leng, and Cheng]{qian2020bggan}
Ming Qian, Congyu Qiao, Jiamin Lin, Zhenyu Guo, Chenghua Li, Cong Leng, and Jian Cheng.
\newblock {BgGAN: Bokeh-glass generative adversarial network for rendering realistic bokeh}.
\newblock In \emph{ECCVW}. Springer, 2020.

\bibitem[Ranftl et~al.(2020)Ranftl, Lasinger, Hafner, Schindler, and Koltun]{ranftl2020towards}
Ren{\'e} Ranftl, Katrin Lasinger, David Hafner, Konrad Schindler, and Vladlen Koltun.
\newblock Towards robust monocular depth estimation: Mixing datasets for zero-shot cross-dataset transfer.
\newblock \emph{IEEE TPAMI}, 44\penalty0 (3), 2020.

\bibitem[Schonberger and Frahm(2016)]{schonberger2016structure}
Johannes~L Schonberger and Jan-Michael Frahm.
\newblock {Structure-from-motion revisited}.
\newblock In \emph{CVPR}, 2016.

\bibitem[Sch{\"o}nberger et~al.(2016)Sch{\"o}nberger, Zheng, Frahm, and Pollefeys]{schonberger2016pixelwise}
Johannes~L Sch{\"o}nberger, Enliang Zheng, Jan-Michael Frahm, and Marc Pollefeys.
\newblock {Pixelwise view selection for unstructured multi-view stereo}.
\newblock In \emph{ECCV}. Springer, 2016.

\bibitem[Shen et~al.(2023)Shen, Li, Sun, Peng, Xian, Cao, and Lin]{shen2023make}
Liao Shen, Xingyi Li, Huiqiang Sun, Juewen Peng, Ke Xian, Zhiguo Cao, and Guosheng Lin.
\newblock Make-it-4d: Synthesizing a consistent long-term dynamic scene video from a single image.
\newblock In \emph{Proceedings of the 31st ACM International Conference on Multimedia}, pages 8167--8175, 2023.

\bibitem[Sheng et~al.(2024)Sheng, Yu, Ling, Cao, Zhang, Lu, Xian, Lin, and Benes]{sheng2024dr}
Yichen Sheng, Zixun Yu, Lu Ling, Zhiwen Cao, Xuaner Zhang, Xin Lu, Ke Xian, Haiting Lin, and Bedrich Benes.
\newblock {Dr. Bokeh: DiffeRentiable Occlusion-aware Bokeh Rendering}.
\newblock In \emph{CVPR}, 2024.

\bibitem[Sun et~al.(2024)Sun, Li, Shen, Ye, Xian, and Cao]{sun2024dyblurf}
Huiqiang Sun, Xingyi Li, Liao Shen, Xinyi Ye, Ke Xian, and Zhiguo Cao.
\newblock {DyBluRF: Dynamic Neural Radiance Fields from Blurry Monocular Video}.
\newblock In \emph{CVPR}, 2024.

\bibitem[Wadhwa et~al.(2018)Wadhwa, Garg, Jacobs, Feldman, Kanazawa, Carroll, Movshovitz-Attias, Barron, Pritch, and Levoy]{wadhwa2018synthetic}
Neal Wadhwa, Rahul Garg, David~E Jacobs, Bryan~E Feldman, Nori Kanazawa, Robert Carroll, Yair Movshovitz-Attias, Jonathan~T Barron, Yael Pritch, and Marc Levoy.
\newblock {Synthetic depth-of-field with a single-camera mobile phone}.
\newblock \emph{ACM TOG}, 37\penalty0 (4), 2018.

\bibitem[Wang et~al.(2023{\natexlab{a}})Wang, Wolski, Pan, Leimk{\"u}hler, Chen, Theobalt, Myszkowski, Seidel, and Serrano]{wang2023implicit}
Chao Wang, Krzysztof Wolski, Xingang Pan, Thomas Leimk{\"u}hler, Bin Chen, Christian Theobalt, Karol Myszkowski, Hans-Peter Seidel, and Ana Serrano.
\newblock {An implicit neural representation for the image stack: Depth, all in focus, and high dynamic range}.
\newblock Technical report, 2023{\natexlab{a}}.

\bibitem[Wang et~al.(2024{\natexlab{a}})Wang, Wolski, Kerbl, Serrano, Bemana, Seidel, Myszkowski, and Leimk{\"u}hler]{wang2024cinematic}
Chao Wang, Krzysztof Wolski, Bernhard Kerbl, Ana Serrano, Mojtaba Bemana, Hans-Peter Seidel, Karol Myszkowski, and Thomas Leimk{\"u}hler.
\newblock {Cinematic Gaussians: Real-Time HDR Radiance Fields with Depth of Field}.
\newblock \emph{arXiv preprint arXiv:2406.07329}, 2024{\natexlab{a}}.

\bibitem[Wang et~al.(2018)Wang, Shen, Zhang, Wang, Lin, Hsieh, Kong, and Lu]{wang2018deeplens}
Lijun Wang, Xiaohui Shen, Jianming Zhang, Oliver Wang, Zhe Lin, Chih-Yao Hsieh, Sarah Kong, and Huchuan Lu.
\newblock {Deeplens: Shallow depth of field from a single image}.
\newblock \emph{arXiv preprint arXiv:1810.08100}, 2018.

\bibitem[Wang et~al.(2023{\natexlab{b}})Wang, Zhao, Ma, and Liu]{wang2023bad}
Peng Wang, Lingzhe Zhao, Ruijie Ma, and Peidong Liu.
\newblock {BAD-NeRF: Bundle adjusted deblur neural radiance fields}.
\newblock In \emph{CVPR}, 2023{\natexlab{b}}.

\bibitem[Wang et~al.(2022)Wang, Yang, Hu, and Zhang]{wang2022nerfocus}
Yinhuai Wang, Shuzhou Yang, Yujie Hu, and Jian Zhang.
\newblock {NeRFocus: Neural radiance field for 3D synthetic defocus}.
\newblock \emph{arXiv preprint arXiv:2203.05189}, 2022.

\bibitem[Wang et~al.(2024{\natexlab{b}})Wang, Chakravarthula, and Chen]{wang2024dof}
Yujie Wang, Praneeth Chakravarthula, and Baoquan Chen.
\newblock {DOF-GS: Adjustable Depth-of-Field 3D Gaussian Splatting for Refocusing, Defocus Rendering and Blur Removal}.
\newblock \emph{arXiv preprint arXiv:2405.17351}, 2024{\natexlab{b}}.

\bibitem[Wang et~al.(2004)Wang, Bovik, Sheikh, and Simoncelli]{wang2004image}
Zhou Wang, Alan~C Bovik, Hamid~R Sheikh, and Eero~P Simoncelli.
\newblock Image quality assessment: from error visibility to structural similarity.
\newblock \emph{IEEE TIP}, 13\penalty0 (4), 2004.

\bibitem[Wei et~al.(2021)Wei, Liu, Rao, Zhao, Lu, and Zhou]{wei2021nerfingmvs}
Yi Wei, Shaohui Liu, Yongming Rao, Wang Zhao, Jiwen Lu, and Jie Zhou.
\newblock Nerfingmvs: Guided optimization of neural radiance fields for indoor multi-view stereo.
\newblock In \emph{ICCV}, 2021.

\bibitem[Wu et~al.(2024)Wu, Yi, Fang, Xie, Zhang, Wei, Liu, Tian, and Wang]{wu20244d}
Guanjun Wu, Taoran Yi, Jiemin Fang, Lingxi Xie, Xiaopeng Zhang, Wei Wei, Wenyu Liu, Qi Tian, and Xinggang Wang.
\newblock {4d gaussian splatting for real-time dynamic scene rendering}.
\newblock In \emph{CVPR}, 2024.

\bibitem[Wu et~al.(2022)Wu, Li, Peng, Lu, Cao, and Zhong]{wu2022dof}
Zijin Wu, Xingyi Li, Juewen Peng, Hao Lu, Zhiguo Cao, and Weicai Zhong.
\newblock {DoF-NeRF}: Depth-of-field meets neural radiance fields.
\newblock In \emph{ACM MM}, 2022.

\bibitem[Xiao et~al.(2018)Xiao, Kaplanyan, Fix, Chapman, and Lanman]{xiao2018deepfocus}
Lei Xiao, Anton Kaplanyan, Alexander Fix, Matt Chapman, and Douglas Lanman.
\newblock {Deepfocus: Learned image synthesis for computational display}.
\newblock In \emph{ACM SIGGRAPH 2018 Talks}. 2018.

\bibitem[Xu et~al.(2024)Xu, Peng, Wang, Blum, Barath, Geiger, and Pollefeys]{xu2024depthsplat}
Haofei Xu, Songyou Peng, Fangjinhua Wang, Hermann Blum, Daniel Barath, Andreas Geiger, and Marc Pollefeys.
\newblock Depthsplat: Connecting gaussian splatting and depth.
\newblock \emph{arXiv preprint arXiv:2410.13862}, 2024.

\bibitem[Yu et~al.(2024)Yu, Chen, Huang, Sattler, and Geiger]{yu2024mip}
Zehao Yu, Anpei Chen, Binbin Huang, Torsten Sattler, and Andreas Geiger.
\newblock {Mip-splatting: Alias-free 3D gaussian splatting}.
\newblock In \emph{CVPR}, 2024.

\bibitem[Zhang et~al.(2024)Zhang, Fang, Shrestha, Liang, Long, and Tan]{zhang2024rade}
Baowen Zhang, Chuan Fang, Rakesh Shrestha, Yixun Liang, Xiaoxiao Long, and Ping Tan.
\newblock {RaDe-GS: Rasterizing Depth in Gaussian Splatting}.
\newblock \emph{arXiv preprint arXiv:2406.01467}, 2024.

\bibitem[Zhang et~al.(2018)Zhang, Isola, Efros, Shechtman, and Wang]{zhang2018unreasonable}
Richard Zhang, Phillip Isola, Alexei~A Efros, Eli Shechtman, and Oliver Wang.
\newblock The unreasonable effectiveness of deep features as a perceptual metric.
\newblock In \emph{CVPR}, 2018.

\bibitem[Zhang et~al.(2019)Zhang, Matzen, Nguyen, Yao, Zhang, and Ng]{zhang2019synthetic}
Xuaner Zhang, Kevin Matzen, Vivien Nguyen, Dillon Yao, You Zhang, and Ren Ng.
\newblock {Synthetic defocus and look-ahead autofocus for casual videography}.
\newblock \emph{arXiv preprint arXiv:1905.06326}, 2019.

\bibitem[Zhao et~al.(2024)Zhao, Wang, and Liu]{zhao2024bad}
Lingzhe Zhao, Peng Wang, and Peidong Liu.
\newblock {BAD-Gaussians: Bundle adjusted deblur Gaussian Splatting}.
\newblock \emph{arXiv preprint arXiv:2403.11831}, 2024.

\end{thebibliography}
}
\clearpage
\setcounter{page}{1}
\maketitlesupplementary

\appendix
To see the dynamic effect of our method and visual comparisons, please refer to our supplementary video. This document includes the following contents:
\begin{itemize}
    \item Details of our synthetic dataset.
    \item Correctness of the proposed dataset.
    \item Color space
    \item Details on all-in-focus experiments.
    \item Details of ablation studies.
    \item Processing time.
    \item Limitations.
\end{itemize}
\section{Details of our synthetic dataset}
\begin{figure}[!t]
  \centering
  \setlength{\belowcaptionskip}{-5pt}
  \includegraphics[width=\linewidth]{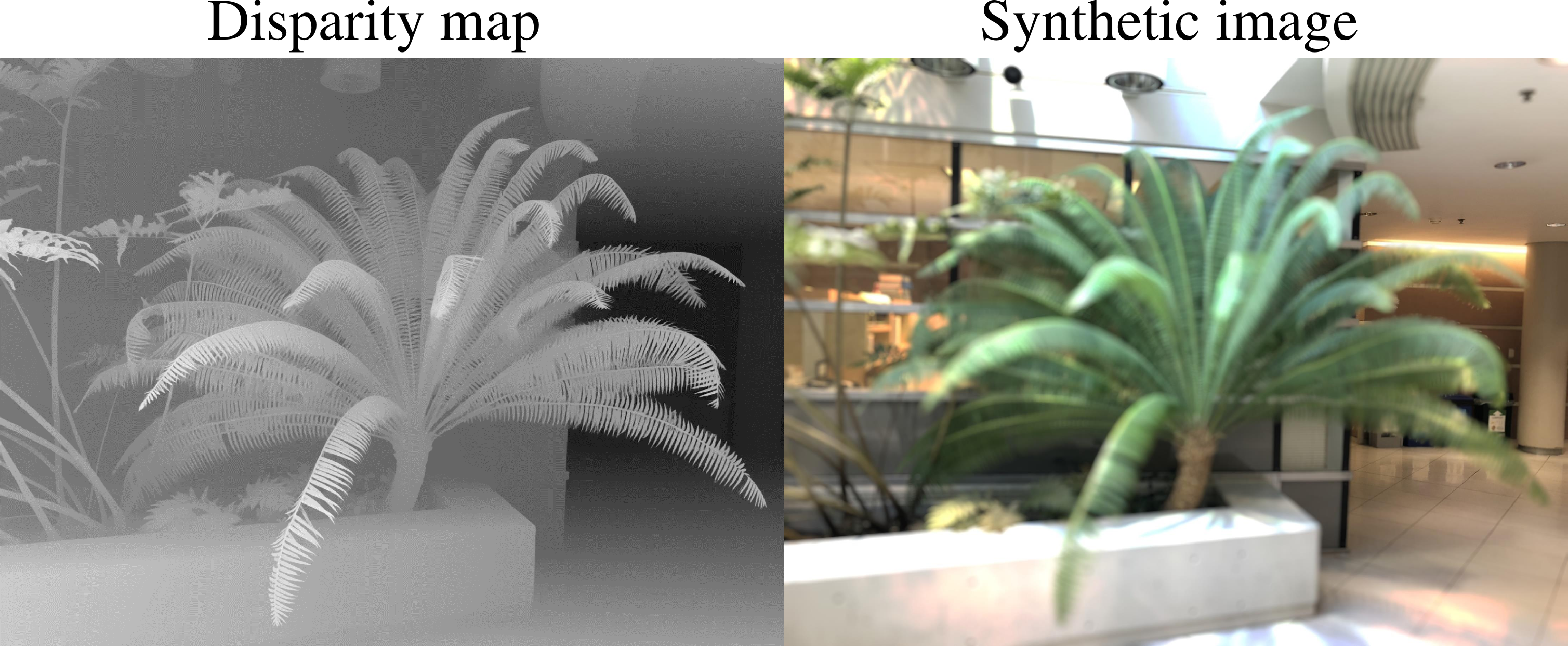}
  \caption{\normalsize{We show the disparity map generated by~\cite{bochkovskii2024depth} and the synthetic shallow DoF image obtained from~\cite{peng2022bokehme}.} 
  }
  \label{fig:supp}
\end{figure}

\begin{figure*}[!t]
  \centering
  \setlength{\belowcaptionskip}{-5pt}
  \includegraphics[width=\linewidth]{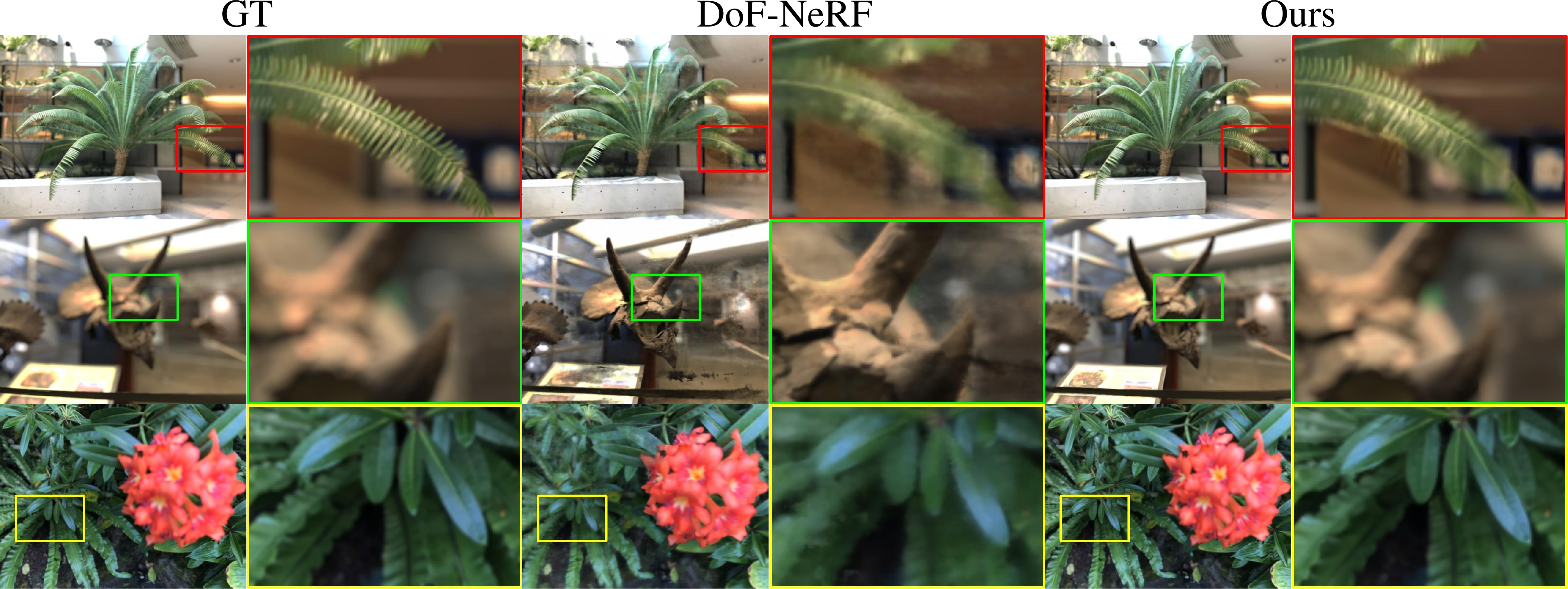}
  \caption{\normalsize{Visual comparison on our synthetic dataset. Our method performs significantly better than DoF-NeRF.} 
  }
  \label{fig:supp2}
\end{figure*}

To quantitatively evaluate the refocusing ability and assess 
whether models learn accurate lens parameters, we introduce a synthetic dataset based on Real Forward Facing dataset~\cite{mildenhall2021nerf} and Tanks and Temples dataset~\cite{kerbl20233d}. Specifically, we apply a state-of-the-art depth estimation method~\cite{bochkovskii2024depth} to generate disparity maps from input images. Subsequently, we employ a single-image DoF rendering method~\cite{peng2022bokehme}, feeding  both the input images and disparity maps into the network to produce images with bokeh blur, as shown in Fig.~\ref{fig:supp}. 
We choose~\cite{peng2022bokehme} to synthesize shallow DoF images primarily because it is predominantly based on traditional physical renderer despite the incorporation of neural networks. The rendered circle of confusion (CoC) in this approach will not be significantly differ from the CoC produced by our lens-based physical imaging model. 
In addition, we excluded Drjohnson and Playroom, two indoor 360° scenes, due to significant monocular depth estimation errors of multi-view input images in indoor environments. At the same time, the inability to generate \textit{poses\_bounds.npy} files for the Train and Truck scenes prevents the evaluation of DoF-NeRF on these two scenes. We maintain these two scenes for comparisons with future 3D-GS methods.
To assess whether the model learns the exact aperture size $\mathcal{A}$ and focus distance$\mathcal{F}$ for each input image, we set these parameters artificially in advance. For the focus distance we set three cases, $\mathcal{F} = 0.2$, $\mathcal{F} = 0.5$ and $\mathcal{F} = 0.8$, corresponding to focus on the background, mid-ground, and foreground, respectively. Recognizing that the aperture size is closely related to the image resolution, we here normalize it to $0-1$ to facilitate the calculation of the error. We consider two cases for aperture size: $\mathcal{A}=0.5$ and $\mathcal{A}=1$.
When we have optimized the 3D-GS scene, we get the learned focus distance and aperture size for each training image. Now, we can we can calculate the lens parameter error as:
\begin{equation}
    \delta_{\mathcal{A}} = \sum_i^N\frac{1}{N}|\mathcal{A}_i-\mathcal{\hat{A}}_i|,
\end{equation}
where $\mathcal{A}$ and $\mathcal{\hat{A}}$ indicate the preset aperture size and learned aperture size, respectively, and N means the number of training images. The smaller this error $\delta_{\mathcal{A}}$ is, the more accurate our learned aperture size is. Similarly we use the following formula to calculate the focus distance error:
\begin{equation}
    \delta_{\mathcal{F}} = \sum_i^N\frac{1}{N}|\mathcal{F}_i-\mathcal{\hat{F}}_i|,
\end{equation}
where $\mathcal{F}$ and $\mathcal{\hat{F}}$ are the preset focus distance and learned focus distance. We use these two metrics to assess whether the model has learned the correct lens parameters. As demonstrated in Tables~\ref{tab:refocus_supp} and~\ref{tab:refocus_delta}, our method outperforms DoF-NeRF~\cite{wu2022dof} in both refocusing ability and the accurate estimation of lens parameters. Furthermore, as illustrated in Fig.~\ref{fig:supp2}, our method generates novel views that are more faithful to the ground-truth images.

\begin{table}[t]
\centering
\caption{
\normalsize{Detailed comparison of our method and DoF-NeRF~\cite{wu2022dof} on our synthetic dataset.}
}
\resizebox{\columnwidth}{!}{
\begin{tabular}{lcccccc}
\toprule
\multirow{2}{*}{Method} & \multicolumn{3}{c}{DoF-NeRF~\cite{wu2022dof}} & \multicolumn{3}{c}{Ours} \\
&PSNR$\uparrow$ & SSIM$\uparrow$ & LPIPS$\downarrow$ & PSNR$\uparrow$ & SSIM$\uparrow$ & LPIPS$\downarrow$ \\
\midrule
Fern& 24.80& 0.736& 0.200& 28.41& 0.867& 0.098\\
Flower& 27.11& 0.82& 0.213& 29.43& 0.889& 0.070\\
Fortress& 29.78& 0.846& 0.186& 32.22& 0.926& 0.059\\
Horns& 24.23& 0.812& 0.235&27.64 & 0.863& 0.122\\
Orchids& 19.99& 0.608& 0.213& 21.54& 0.659& 0.165 \\
Room& 26.55& 0.842& 0.198& 32.16& 0.933& 0.071\\
Trex& 26.65& 0.853& 0.207& 29.53& 0.910& 0.082\\
Train& ---& ---& ---& 22.71& 0.676& 0.216\\
Truck& ---& ---& ---& 21.09& 0.675& 0.312\\
\bottomrule
\end{tabular}

}
\vspace{-0.05in}
\label{tab:refocus_supp}
\end{table}

Compared to the previous datasets proposed by Ma \etal~\cite{ma2022deblur} and Wu \etal~\cite{wu2022dof}, which evaluate defocus deblurring ability, our dataset is specifically designed to assess refocusing capabilities. Both the training and test sets in our synthetic dataset consist of shallow DoF images. We hope that this dataset will facilitate future work in this field.

\begin{table}[htb]
\centering
\caption{
\normalsize{Detailed comparison of our method and DoF-NeRF~\cite{wu2022dof} on our synthetic dataset.}
}
\resizebox{0.7\columnwidth}{!}{
\begin{tabular}{lcccc}
\toprule
\multirow{2}{*}{Method} & \multicolumn{2}{c}{DoF-NeRF~\cite{wu2022dof}} & \multicolumn{2}{c}{Ours} \\
&$\delta_\mathcal{A}\downarrow$ & $\delta_\mathcal{F}\downarrow$ & $\delta_\mathcal{A}\downarrow$ & $\delta_\mathcal{F}\downarrow$ \\
\midrule
Fern&0.204 &0.263 & 0.089& 0.102\\
Flower&0.127 &0.280 &0.091 &0.074 \\
Fortress&0.156 &0.299 & 0.187& 0.021\\
Horns& 0.234&0.205 &0.197 &0.075 \\
Orchids& 0.189&0.219 & 0.097&0.087  \\
Room& 0.276&0.278 &0.066 &0.116 \\
Trex& 0.189&0.251 &0.154 &0.079 \\
Train& ---& ---&0.225 &0.113 \\
Truck& ---& ---&0.258 &0.148 \\
\bottomrule
\end{tabular}

}
\vspace{-0.05in}
\label{tab:refocus_delta}
\end{table}
\section{Correctness of the proposed dataset}
We validate the accuracy of the synthesis strategy using the BLB dataset, which comprises 500 test samples, each containing paired all-in-focus and defocus images. All-in-focus images are processed through our synthesis pipeline, and the resulting synthesized defocus images are compared with the ground truth to calculate PSNR and SSIM metrics. The High PSNR and SSIM values indicate that the synthesized bokeh is close to the real, thereby confirming the effectiveness of our synthesis strategy.
\begin{table}[h]
\centering
\caption{
\normalsize{The High PSNR and SSIM values indicate that our synthesized bokeh is close to the real.}
}
    \begin{tabular}{lcc}
\toprule
    & PSNR$\uparrow$  & SSIM$\uparrow$ \\ 
    \midrule
    Ours &  43.30 & 0.9932 \\
    \bottomrule
\end{tabular}
\end{table}
\section{Color space.}
We apply a gamma transform on the input image to convert it from sRGB color space to linear color space. Subsequently, we simulate the circle-of-confusion within the linear color space. Finally, gamma correction is performed to convert the image from linear space back to sRGB space. The gamma value is $2.2$. This process will be further emphasized in our revised version.

\section{Details on all-in-focus experiments}

As shown in Table~\ref{tab:clear}, we present the per-scene breakdown results of Real Forward-facing~\cite{mildenhall2021nerf} and T\&T\_DB~\cite{kerbl20233d} datasets. These results align with the averaged results presented in the main text. Our method is built upon Mip-Splatting~\cite{yu2024mip}, a robust 3D-GS approach for all-in-focus inputs. Evidently, our method demonstrates superior performance compared to Mip-Splatting in most scenes.
This indicates that our method can not only handle shallow DoF inputs, but also performs excellent under general input conditions, specially on Real Forward-facing dataset. 

\begin{table}[htb]
\centering
\caption{
\normalsize{Detailed comparison of other methods and ours on the all-in-focus dataset.}
}
\resizebox{\columnwidth}{!}{
\begin{tabular}{lcccccc}
\toprule
\multirow{2}{*}{Method} & \multicolumn{3}{c}{Mip-Splatting~\cite{yu2024mip}} & \multicolumn{3}{c}{Ours} \\
&PSNR$\uparrow$ & SSIM$\uparrow$ & LPIPS$\downarrow$ & PSNR$\uparrow$ & SSIM$\uparrow$ & LPIPS$\downarrow$ \\
\midrule
Fern& 27.87&  0.910 & 0.062 & \textbf{28.23} & \textbf{0.917} & \textbf{0.062}\\
Flower& 25.55& 0.868& 0.101& \textbf{27.81}& \textbf{0.910}&\textbf{0.059}\\
Fortress& 27.85& 0.913& 0.072& \textbf{32.57}& \textbf{0.951}& \textbf{0.032}\\
Horns& 29.47 & 0.952& \textbf{0.049}& \textbf{30.29}& \textbf{0.954}&0.052\\
Leaves& \textbf{22.42}& 0.848& 0.091& 22.36& \textbf{0.850}&\textbf{0.089}\\
Orchids& 21.58& 0.800& 0.105& \textbf{22.10}& \textbf{0.821}&\textbf{0.090} \\
Room& 33.92& 0.963& 0.057& \textbf{34.63}& \textbf{0.973}& \textbf{0.052}\\
Trex& 28.67& 0.951& 0.056& \textbf{30.29}& \textbf{0.961}&\textbf{0.042}\\
Train& \textbf{21.89}& \textbf{0.821}& \textbf{0.190}& 21.15&  0.791& 0.251\\
Truck& \textbf{25.43}& \textbf{0.888}& \textbf{0.129}& 24.61& 0.872& 0.180\\
Playroom& 30.50& 0.916& \textbf{0.223}& \textbf{30.93}& \textbf{0.924}& 0.242\\
Drjohnson& \textbf{29.44}& 0.890& \textbf{0.249}& 29.37& \textbf{0.895}& 0.258\\
\bottomrule
\end{tabular}

}
\vspace{-0.05in}
\label{tab:clear}
\end{table}

% Fern& 28.50&  0.923 & 0.048 \\
% Flower& 27.28& 0.891& 0.053&\\
% Fortress& 30.70& 0.903& 0.039\\
% Horns& 29.81 & 0.949& 0.049\\
% Leaves&21.77& 0.839& 0.077\\
% Orchids& 21.73& 0.811& 0.073\\

\section{Details of ablation studies}
\begin{figure*}[!t]
  \centering
  \setlength{\belowcaptionskip}{-5pt}
  \includegraphics[width=\linewidth]{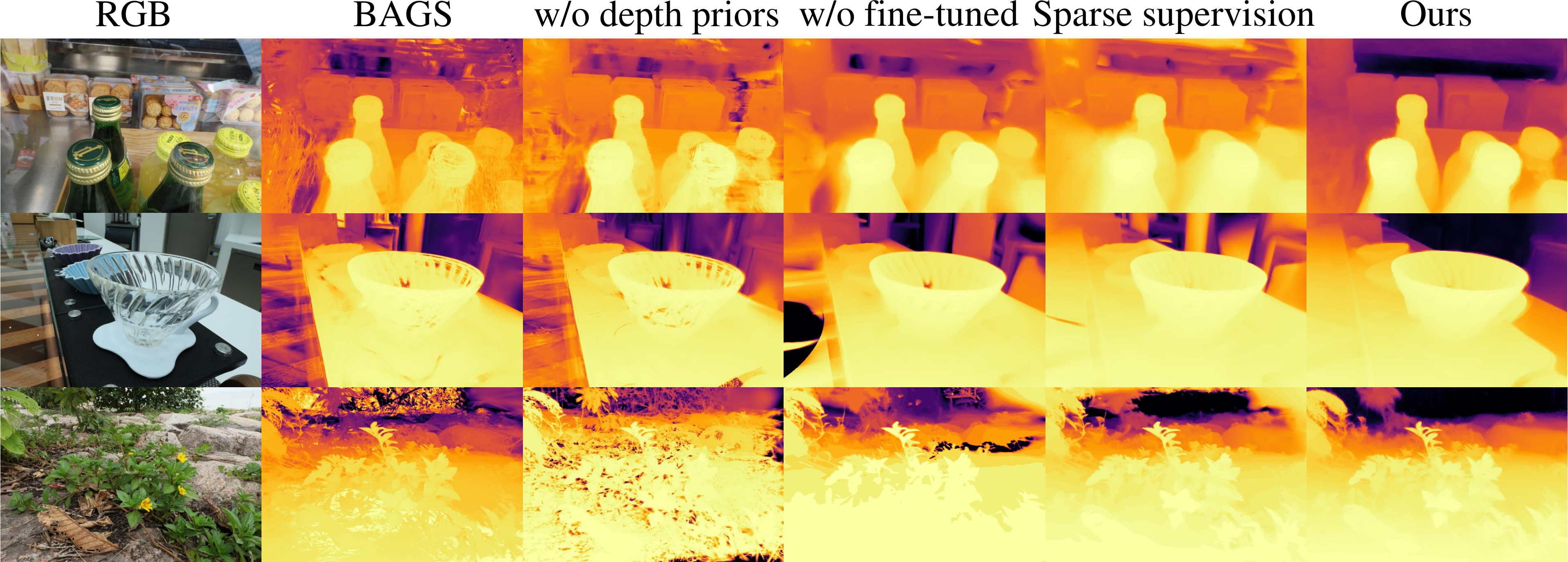}
  \caption{\normalsize{Visual comparison of different depth supervision strategies.} 
  }
  \label{fig:supp1}
\end{figure*}
\begin{table}[h]
\centering
\caption{
\normalsize{comparisons on processing time.}
}
\resizebox{\columnwidth}{!}{
\begin{tabular}{lccccc}
\toprule
Method & Deblur-NeRF~\cite{ma2022deblur}& DoF-NeRF~\cite{wu2022dof} & BAGS~\cite{peng2024bags} & Deblurring 3DGS~\cite{lee2024deblurring} & \textbf{Ours} \\
\midrule
% BAGS~\cite{peng2024bags}& 23.91& 0.704& 0.293&--- &---\\
Time& 20 hours& 11 hours& 25 mins& 10 mins& 18 mins \\
FPS& $<1$&$<1$& 332& 381& 364  \\
\bottomrule
\end{tabular}

}
\vspace{-0.05in}
\label{tab:time}
\end{table}
\begin{table*}[htbp]
\centering
\caption{
\normalsize{Ablation studies of per-scene breakdown results on the defocus deblurring dataset~\cite{ma2022deblur}.}
}
\resizebox{\linewidth}{!}{
\begin{tabular}{lcccccccccccccccccc}
\toprule
\multirow{2}{*}{Method}& \multicolumn{3}{c}{baseline} & \multicolumn{3}{c}{w/o lens} & \multicolumn{3}{c}{w/o depth} & \multicolumn{3}{c}{w/o adaptation} &\multicolumn{3}{c}{sparse depth} & \multicolumn{3}{c}{w/o fine-tuned depth}\\
&PSNR & SSIM & LPIPS & PSNR & SSIM & LPIPS& PSNR & SSIM & LPIPS& PSNR & SSIM & LPIPS& PSNR & SSIM & LPIPS& PSNR & SSIM & LPIPS \\
\midrule
cake&24.15& 0.710& 0.216& 25.69&  0.782 & 0.131 & 26.51 & 0.800 & 0.116& 26.09& 0.794& 0.107 & 26.43& 0.795& 0.121& 26.41& 0.792& 0.133\\
caps&21.24& 0.559& 0.332& 23.57& 0.713& 0.148& 24.41& 0.741& 0.145& 24.12& 0.737& 0.142& 24.52& 0.742& 0.164& 24.52 & 0.745& 0.157\\
cisco&20.77& 0.732& 0.114&  20.85& 0.743& 0.069& 20.95& 0.742& 0.071& 20.88& 0.739& 0.067& 20.72& 0.734& 0.079& 20.76& 0.736& 0.082\\
coral& 19.66& 0.568& 0.288& 19.51&0.599& 0.147& 19.86& 0.608& 0.122& 19.71& 0.602& 0.133& 19.87& 0.605& 0.132& 19.89& 0.603& 0.132\\
cupcake&21.72& 0.686& 0.198& 22.09& 0.742& 0.089& 22.82& 0.757& 0.079& 22.63& 0.752& 0.080& 22.74& 0.752& 0.0087& 22.81& 0.752& 0.086\\
cups&24.29& 0.749& 0.223& 25.89& 0.814& 0.100& 25.91& 0.818& 0.114& 26.06& 0.820& 0.086& 25.34& 0.800& 0.115& 25.63& 0.804& 0.117 \\
daisy&18.00& 0.493&0.299& 23.35& 0.734& 0.062& 23.33& 0.721& 0.086& 23.54& 0.724& 0.069& 22.88& 0.706& 0.114& 22.80& 0.700& 0.119\\
sausage&17.45& 0.461& 0.284& 17.99& 0.515& 0.169& 18.47& 0.536& 0.151& 18.29& 0.529& 0.156& 18.18& 0.531& 0.172& 18.55& 0.550& 0.153\\
seal& 20.71& 0.561& 0.288& 24.34& 0.744& 0.114& 25.54& 0.790& 0.105& 25.34& 0.781& 0.088& 26.17& 0.805& 0.095& 26.10& 0.804& 0.097\\
tools& 25.09& 0.845& 0.152& 27.17& 0.898& 0.056& 28.09& 0.911& 0.051& 27.53& 0.902& 0.052& 27.57& 0.900& 0.061& 27.82& 0.902& 0.059\\
\bottomrule
\end{tabular}

}
\vspace{-0.05in}
\label{tab:ablation_all}
\end{table*}
In this section, we present detailed results of the ablation experiments in our main paper.
In Table~\ref{tab:ablation_all}, we show the per-scene breakdown results of the ablation studies---baseline, w/o lens-based imaging model, w/o per-scene depth priors,  w/o defocus-to-focus adaptation, sparse depth supervision, and no fine-tuned depth supervision. This indicates that each component of our system plays an important role in improving the image deblurring quality. In addition, we demonstrate the effectiveness of our approach by showing a visual comparison of the depth maps rendered by 3D-GS under different depth strategies, as shown in Fig.~\ref{fig:supp1}. 

\section{Processing time.}
As shown in Table~\ref{tab:time}, we recorded the processing time for both our method and other approaches on a single NVIDIA RTX A6000 GPU. For both Deblur-NeRF~\cite{ma2022deblur} and DoF-NeRF~\cite{wu2022dof}, we follow the specified training iterations outlined in the original papers, and calculate the training time. Due to the underlying NeRF-based framework, their average training time on the defocus deblurring dataset~\cite{ma2022deblur} is approximately 20 hours and 11 hours, respectively. Furthermore, their inference time is observed to be notably slow, achieving frame rates below 1 FPS. For the 3D-GS methods---BAGS~\cite{peng2024bags}, Deblurring 3DGS~\cite{lee2024deblurring}, and our method, we uniformly train for 30k iterations and record the training time and FPS. Although our method incorporates a lens imaging model, the training time is only slightly affected and it remains faster than BAGS. Benefit from the 3DGS framework, all GS-based methods can achieve fast rendering, obtaining FPS of approximately 360. In addition to training 3D Gaussian Splatting model, it takes about 3 minutes to fine-tune the depth network for per-scene depth priors.

\section{Discussion on depth supervision.}
We employ per-scene adjustments of depth priors to guide the reconstruction and ensure the accurate scene geometry and rendered depth maps. The effectiveness of this approach is demonstrated by ablation experiments. However, depth maps predicted by the fine-tuned depth network are not entirely accurate, and using these as pseudo-gt to supervise the depth maps rendered by 3D-GS introduces a degree of noise. This residual noise may impact the precision of the final depth maps, particularly in scenes with complex geometry. We therefore use a strategy of gradual decay of the depth loss weight $w_d$. In particular, we gradually decay this weight to $1/10$ of the initial value.

\section{Limitations}
Our method may encounter limitations when the blur is view-consistent, such as in cases where the camera maintains a fixed focal point, \ie., focusing on a single target). Specifically, when the multi-view inputs all focus on the foreground, our method may struggle to recover clear background information. Consequently, a sharp scene can only be reconstructed if the input images contain both focused foreground and focused background elements.
Addressing defocus deblurring under view-consistent conditions may be feasible through the integration of image priors, which we consider as a direction for future work.


\end{document}